\documentclass[journal]{IEEEtran}
\usepackage{xcolor}

\usepackage{capekColors}
\usepackage{capekCommands}

\usepackage{algpseudocode}
\usepackage{algorithm}
\usepackage{booktabs}
\usepackage{multirow}
\usepackage{siunitx}
\usepackage{footnote}
\pgfplotsset{compat=1.18}
\makesavenoteenv{tabular}

\usepackage[]{hyperref}
\hypersetup{
    colorlinks,
    linkcolor={PairedF!70!black}, 
    citecolor={PairedB!70!black},
    urlcolor={PairedB!70!black}
}

\usepackage{tikz}
\usetikzlibrary{patterns, arrows, arrows.meta, arrows.spaced, calc, angles, quotes}

\usepackage{graphicx}
\graphicspath{{figures/}}

\usepackage[precision=2, unit=mm]{lengthconvert}

\newcommand{\gene}{\M{g}}
\newcommand{\weight}{\M{w}}
\newcommand{\Dmat}{\M{D}}

\begin{document}
\title{Multi-objective Memetic Algorithm with Adaptive Weights for Inverse Antenna Design}
\author{Petr~Kadlec, \IEEEmembership{Member, IEEE} and Miloslav~Capek, \IEEEmembership{Senior Member, IEEE}
\thanks{P. Kadlec is with the Brno University of Technology, Brno, Czech Republic (e-mail: kadlecp@vut.cz).}
\thanks{M. Capek is with the Czech Technical University in Prague, Prague, Czech Republic (e-mail: miloslav.capek@fel.cvut.cz).}
}

\maketitle

\begin{abstract}
This paper \RR{deals with discrete topology optimization and} describes the modification of a single-objective algorithm into its multi-objective counterpart. \RR{The result is a significant increase in the optimization speed and quality of } the resulting Pareto front as compared to conventional state-of-the-art automated inverse design \RR{techniques}. This advancement is possible thanks to a memetic algorithm combining a gradient-based search for local minima with heuristic optimization to maintain sufficient diversity. The local algorithm is based on rank-1 perturbations; the global algorithm is NSGA-II. An important advancement is the adaptive weighting of objective functions during optimization. The procedure is tested on \RR{four} challenging examples dealing with both physical and topological metrics and multi-objective settings. The results are compared with standard techniques, and the superb performance of the proposed technique is reported. The implemented algorithm applies to antenna inverse design problems and is an efficient data miner for machine learning tools.
\end{abstract}


\section{Introduction}
\label{sec:intro}

Antenna design is an immensely complex problem involving the determination of optimal shape, material distribution, and excitation. All these aspects have to be taken into account together. In addition, many constraints are commonly applied, \eg{}, the space available for the design, the close surroundings of an antenna, or the technological process used~\cite{Balanis_Wiley_2005, Fujimoto_Morishita_ModernSmallAntennas, VolakisChenFujimoto_SmallAntennas}. 

Powerful tools were developed to assist the design process~\cite{KozielOgurtsov_AntennaDesignBySimulationDrivenOptimization, Simon_EvolutionaryOptimizationAlgorithms, BendsoeSigmund_TopologyOptimization, Ohsaki_OptimizationOfFiniteDimensionalStructures} and are frequently utilized by the community, the most prominent being topology optimization~\cite{BendsoeSigmund_TopologyOptimization, JohnsonSamii1997_GAinEM, Haupt_Werner_GeneticAlgorithmsInEM} which is robust and capable of finding unconventional shapes. The classical heuristic methods~\cite{Simon_EvolutionaryOptimizationAlgorithms}, often powered by genetic algorithms (GA)~\cite{JorhsonRahmatSamii_GAandMoMforTheDesignOfIntegratedAntennas, Chen2016, CismasuGustafsson_FBWbySimpleFreuqSimulation}, are robust but relatively slow due to their global nature. In contrast, methods such as adjoint formulation~\cite{HassanWadbroBerggren_TopologyOptimizationOfMetallicAntennas, 2016_Liu_AMS, Tucek_etal_DensityBasedTOinMoMforQmin_2023} are fast, although their rapid convergence to the local minima is related to a specific set of regularization filters and might fail in some cases, especially when considering self-penalized scenarios~\cite{TucekEtAl_EuCAP2024_TopoOptGain}. A discrete, gradient-based approach~\cite{Capeketal_ShapeSynthesisBasedOnTopologySensitivity, Jiang_etal_PixelAntenna_2022, WangHum_APS_BinaryTopologyModel} might be an alternative to the adjoint formulation. However, it becomes computationally expensive for many binary optimization unknowns~\cite{2021_capeketal_TSGAmemetics_Part2}. Recently, global and local approaches have been combined into a fast and robust memetic scheme~\cite{2021_capeketal_TSGAmemetics_Part1}, thus eliminating the above-mentioned disadvantages.

Nevertheless, each engineering problem is implicitly multi-objective~\cite{Deb_MultiOOusingEA}, \ie{}, the trade-off between physical performance, cost, and manufacturability must be considered. \RR{Handling multiple objectives is challenging because the set of optimal trade-offs, known as the Pareto front, needs to be accurately represented to ensure \RR{well-informed} decision-making~\cite{Deb_MultiOOusingEA}.} 

Multi-objective problems have commonly been solved by accumulating all objectives \RR{multiplied by individual weighting coefficients} into a composite objective function~\cite{Ehrgott_MulticriteriaOptimization, Deb_MultiOOusingEA} and then running the optimization multiple times with different sets of weights. The process of reducing a multi-objective (MO) problem into a series of single-objective (SO) problems, called scalarization, prioritizes different objectives but takes a long time and is suboptimal as the variation of the weights is defined \textit{a priori}. Consequently, the resulting Pareto front is of low quality, \ie{}, many points are dominated, the coverage of the Pareto front is non-uniform, and/or data are missing for a feasible combination of objectives. \RR{In this paper, the approach when a single-objective GA-based optimizer with \textit{a priori} chosen weights is iteratively restarted for different weight combinations is denoted by the abbreviation SOGA-FW.} 

Some of the drawbacks of scalarization, \eg{}, the inability to find non-convex parts of the Pareto front, can be solved by advanced weighting methods such as the Rotated Weighted Metric method, or $\epsilon$-constrained method~\cite[Ch.~3]{Deb_MultiOOusingEA}. \RR{However, a key drawback remains: the insights gained from solving one single-objective problem are not carried over when solving another one, thus, reducing overall efficiency.} The idea of combining agents with different weighting vectors to solve more SO problems simultaneously was introduced in \cite{hajela1992genetic}. There, an \RR{auxiliary} integer variable selecting the weighting vector from a set of \textit{a priori} defined vectors is added to the list of variables of every agent. Consequently, a new multi-objective GA was proposed~\cite{murata1995moga}, selecting weights randomly for every proposed genome and keeping the non-dominated solutions in the so-called external archive~\cite{Deb_MultiOOusingEA}. An MO GA-based algorithm performing a local search on neighborhood agents with Hamming distance one was then introduced in~\cite{ishibuchi1998multi}. Recent studies, \eg{},~\cite{parraga2017using, ma2019nsga, leung2020hybrid} have shown that \RR{hybridization of well-used optimization methods, such as the Elitist Non-dominated Sorting Genetic Algorithm (NSGA-II)~\cite{deb2002fast} or Multi-Objective Evolutionary Algorithm based on Decomposition (MOEA/D)} \cite{trivedi2016survey}, with local search, significantly improves performance. Building on this, \cite{junqueira2022multi} proposed an efficient method to adaptively change individual agent weights based on their positions relative to the currently used set of weighting vectors.

\RR{This paper describes how adaptive weights are incorporated into the memetic algorithm~\cite{2021_capeketal_TSGAmemetics_Part1, 2021_capeketal_TSGAmemetics_Part2}, enabling it to search for Pareto fronts efficiently.} Each agent operating within the heuristic algorithm is equipped with its individual set of weights which evolve during the optimization. The key enabler for this modification is the local gradient-based routine, which moves every agent quickly into the local minimum~\cite{2021_capeketal_TSGAmemetics_Part1}. Hence, the heuristic operates in the subspace of local minima only, and the entire MO problem is reduced in determining the optimal weights to get the best approximation of the Pareto front. Different weights associated with the agents add extra diversity to the existing procedure, improving the desired robustness to solve high-dimensional non-linear problems. 

The main advantages over existing methods are that the optimization problem is solved only once (\RR{the whole process is approximately ten to hundred-times faster compared to the scalarization technique}), and the quality of the Pareto front is considerably higher. The fact that agents of different priorities, \ie{}, having different MO weights, coexist within the same optimization run provides the algorithm with additional robustness and saves computational time concurrently.

The paper is organized as follows. The optimization procedure is introduced in Section~\ref{sec:method} where the memetic algorithm is reviewed and the technique to determine the weights adaptively is presented. \RR{Four} examples are shown in Section~\ref{sec:examples}, demonstrating the efficacy of the method. Possible extensions are discussed in Section~\ref{sec:disc}, and the paper is concluded in Section~\ref{sec:conclu}. Important implementation details are elaborated in a series of Appendices. The paper is accompanied by supplementary material~\cite{KadlecCapek2024}, containing detailed results of the optimization problems elaborated upon in Section~\ref{sec:examples}.

\section{Adaptive Weights in a Memetic Algorithm}
\label{sec:method}

The effective evaluation of MO optimization is enabled by combining \RR{a local, gradient-based, and a global, usually heuristic, algorithms~\cite{Simon_EvolutionaryOptimizationAlgorithms}. A gradient-based algorithm detects local minima quickly, and the heuristic algorithm maintains diversity among searched candidate solutions.} Such a combination is generally called a memetic algorithm~\cite{Hart_etal_RecentAdvancesInMemeticAlgorithms}. Here, we utilize the recently proposed technique~\cite{2021_capeketal_TSGAmemetics_Part1} based on the rank-1 updates of the impedance matrix \RR{associated with all additions or removals of optimization degrees of freedom. The final structural update in each iteration follows the steepest descent in the objective function. The efficacy of the SO procedure} was reported in~\cite{2021_capeketal_TSGAmemetics_Part2}. The essential parts of this algorithm are briefly recapitulated in Section~\ref{sec:meme}, and its MO modification based on variable weights is described in Sections~\ref{sec:newWeights} and \ref{sec:asignWeights}.

\subsection{Single-Objective Memetic Algorithm}
\label{sec:meme}

\begin{figure}
    \centering
    \includegraphics[width=\columnwidth]{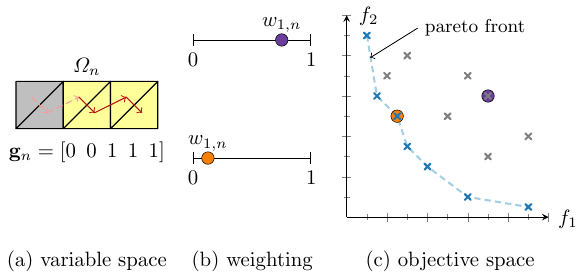}
    \caption{Three parameter spaces used in this paper. (a) The design region is discretized, here into a set of triangles, and antenna~$\srcRegion_n$ is represented by vector~$\gene_n$, with zeros corresponding to a vacuum and ones corresponding to a metal. The values of vector~$\gene_n$ form the variable space. (b) The weighting sum method is used in this work to deal with multi-objective (MO) problems~\cite{Ehrgott_MulticriteriaOptimization}. The vector of weights~$\M{w}$ is assigned from a normalized barycentric coordinate system. (c) Antenna performance is studied in the objective space. The variables (shapes) and weights are set so that the Pareto front demonstrates the trade-off between optimized parameters.}
    \label{fig1}
\end{figure}

The shape of antenna~$\srcRegion_n$ is optimized as
\begin{equation}
\begin{aligned}
	& \mathrm{min} && f (\srcRegion_n) \\
	& \mathrm{subject\,\,to} && \srcRegion_n \subseteq \srcRegion_0. \\
\end{aligned}
\label{eq:SOOP1}
\end{equation}
Design region~$\srcRegion_0$ is discretized first and the excitation is fixed. Possible~$\srcRegion_n$ shapes are represented with binary vectors~$\gene_n$, see Fig.~\ref{fig1}. It was shown in~\cite{2021_capeketal_TSGAmemetics_Part1} that the performance of \RR{rank-1} modifications of shape~$\gene_n$ can be \RR{calculated} two orders faster than the regular solution using the method of moments~\cite{Harrington_FieldComputationByMoM}. The repetition of rank-1 modifications readily improves shape~$\srcRegion_n$ so its performance is optimized. The local algorithm usually stops in the local minimum due to the non-convexity of objective function~$f$ (maximum bandwidth, optimal matching, etc.) defining the SO problem. To provide robustness, GA is applied as the global step and uses the local minima found in the previous iteration to propose new shapes to be locally optimized. The algorithm stops after a given number of iterations or when the convergence criteria are met~\cite{2021_capeketal_TSGAmemetics_Part2}.
 
\subsection{Objective Weights as Optimization Unknowns}
\label{sec:weights}

Almost every design problem considers more than one objective. Therefore, a general MO optimization formulation can be applied to any antenna design problem:
\begin{equation}
\begin{aligned}
	& \mathrm{min} && f_m (\srcRegion_n), \quad m \in \left\{1,\dots, M\right\} \\
	& \mathrm{subject\,\,to} && \srcRegion_n \subseteq \srcRegion_0. \\
\end{aligned}
\label{eq:MOOP1}
\end{equation}
Here, $M$ stands for the number of individual objective functions~$f_m$ and~$\srcRegion_n$ denotes the~$n$-th antenna shape, see Fig.~\ref{fig1}, that is defined by \RR{a binary} vector~$\M{g}_n$. Solving the problem~\eqref{eq:MOOP1} properly leads to a Pareto front which describes the optimal trade-off between individual objectives~$f_m$.

The composite objective function is formed by the convex combination of all individual objectives $f_m$ as
\begin{equation}
    \widetilde{f}  = \M{w}^\trans \M{f} \left(\srcRegion_n \right), \quad \sum_m w_m = 1,
    \label{eq:composedFitness}
\end{equation}
where $\M{w} = [w_1, \dots, w_M]^\trans$ is a column vector of weights and $\M{f} (\srcRegion_n) = [f_1(\srcRegion_n), \dots, f_M(\srcRegion_n)]^\trans$ is a vector of individual objective function values.

As the objectives usually have different scaling and/or physical units, they should be normalized before being added up to form a single value using~\eqref{eq:composedFitness}. It is impossible to select a combination of weights that would pick a required point on a Pareto front \textit{a priori} without knowing the shape of the Pareto front\footnote{The scaling of the weights is usually highly non-linear, and working with, \eg{}, the combination of weights $\M{w} = \left[0.5,0.5\right]^\trans$ does not guarantee a solution that would be in the middle section of the Pareto front.}. Consequently, finding a uniform representation of the Pareto front is expensive.

The local algorithm introduced in \cite{2021_capeketal_TSGAmemetics_Part1} excels in the exploitation because of its ability to quickly reach the local minima~\cite{2021_capeketal_TSGAmemetics_Part1} and can, therefore, be applied to MO problems. In this case, many individual runs solving~\eqref{eq:composedFitness} with \textit{a priori} selected combinations of weights~$\M{w}$ are needed. The entire set of GA agents searches for the same local minima~$\widetilde{f}$, corresponding to a particular point on the Pareto front. This is, however, ineffective; different weighting vectors~$\M{w}$ often lead to the same or similar solution represented by~$\M{f}(\srcRegion_n)$. 

\subsection{Adaptive Weighting of Objective Functions}
\label{sec:adaptWeights}
To overcome the issue with a manual repetitive scalarization of the MO problem, a new algorithm combining traditional multi-objective NSGA-II~\cite{deb2002fast} searching for a set of non-dominated solutions with a powerful local algorithm based on topology optimization~\cite{2021_capeketal_TSGAmemetics_Part1} is proposed. Our approach combines the strengths of both algorithms:
\begin{enumerate}
    \item NSGA-II aims for good coverage of the Pareto front and excels in maintaining the diversity among the agents~\cite{deb2002fast},
    \item the local algorithm identifies local minima~\cite{2021_capeketal_TSGAmemetics_Part1}.
\end{enumerate}
\RR{While NSGA-II operates within a MO formulation~\eqref{eq:MOOP1}, \ie{}, evaluating solutions based on a vector of objective values~$\left[f_1, \dots, f_M\right]^\trans$, the local algorithm uses the SO formulation~\eqref{eq:composedFitness}, \ie{}, the single value~$\widetilde{f}$.} Therefore, an effective strategy is needed to assign each NSGA-II agent with its weighting vector. Hence, the agents can seek different parts of the Pareto front which naturally increase their diversity. This addresses a limitation noted in \cite{2021_capeketal_TSGAmemetics_Part1}. Additionally, the quality of the proposed solutions improves as only locally optimal solutions are used to generate new trial solutions. These new trial solutions consist of features that are locally optimal, usually from slightly different perspectives, \ie{}, using different weighting vectors.    

\begin{figure}
\centering
\includegraphics[width=\columnwidth]{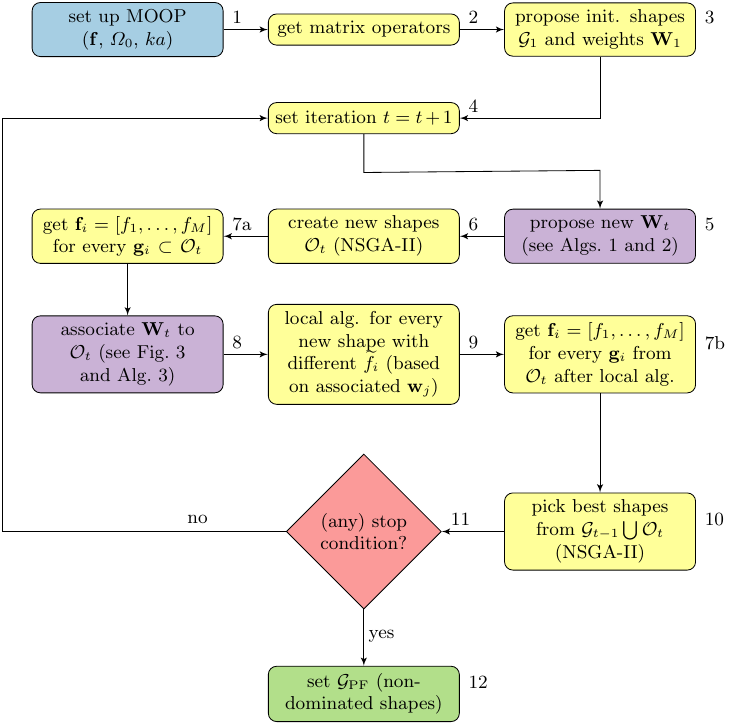}
\caption{The flowchart of the \RR{proposed} multi-objective memetic algorithm with adaptive weights (MOMA-AW) proposed in this paper. Here, $\M{f}$ denotes the vector of objective function values, matrix~$\M{W}$ is the set of weighting vectors, $\mathcal{G}$ and $\mathcal{O}$ represent the sets of agents (generation) and offspring candidates, respectively.}
\label{fig:flowChart}
\end{figure}

Our proposed algorithm is called a Multi-objective Memetic Algorithm with Adaptive Weights (MOMA-AW) and its workflow is illustrated in Fig.~\ref{fig:flowChart}. It starts with a proper definition of the MO problem, \ie{}, the definition of a set of objective functions~$\M{f} = \left[f_1, \dots, f_M \right]^\trans$, the design region $\Omega_0$, and the electrical size of the problem $ka$. The second step is to obtain matrix operators, \eg{}, the impedance matrix, for the given design region. The optimization process starts at iteration~$t = 1$ with a random seed of $N$~initial shapes $\mathcal{G}_1 = \left\{\M{g}_1, \dots,\M{g}_N\right\}$ (the so-called generation) and a user-defined set of $N$~weighting vectors. The weighting vectors and individual objective function values for one generation of agents are stored in the form of matrices: $\M{W} = \left[ \M{w}_1, \dots , \M{w}_N \right]$ and $\M{F} = \left[ \M{f}_1, \dots , \M{f}_N \right]$, respectively. At initial iteration $t = 1$, we distribute the weighting vectors so that they cover an $M$-dimensional simplex uniformly as proposed in \cite{das1998normal}. 

\RR{After algorithm´s initialization steps, the main loop of the global optimization starts with the incrementation of the timer. Then, new weighting vectors are stored in~$\M{W}_t$. The process of updating the weights is described in Section~\ref{sec:newWeights}. After that, a set of offspring shapes~$\mathcal{O}_{t}$ is created by well-known GA strategies, namely crossover and mutation. A standard crossover cutting two randomly chosen parents from the mating pool at a randomly chosen position and combining them to produce two offspring shapes is used. This simplest crossover method is used since there is no correspondence between a geometrical topology of DOFs (\ie{} the basis functions) and the ordering of individual bits in the binary representation of shape $\M{g}$.} These processes are controlled by the probability of crossover~$p_{\mathrm{C}}$ and the probability of mutation~$p_{\mathrm{M}}$, respectively. 

The objective functions are evaluated for all candidate shapes collected in~$\mathcal{O}_{t}$. Based on these values, individual weights from matrix~$\M{W}_t$ are associated with individual shapes from~$\mathcal{O}_{t}$ (see Section~\ref{sec:asignWeights}). The individual composite objective function~$\widetilde{f}$ in~\eqref{eq:composedFitness} is constructed for every new candidate. Then, a local algorithm is called, searching for a local optimum, ideally near the desired area of the Pareto front searched. Finally, the locally optimized shapes~$\mathcal{O}_t$ are combined with the parent population~$\mathcal{G}_{t-1}$.  The new generation~$\mathcal{G}_t$ of size~$N$ is chosen based on NSGA-II procedures, \ie{}, non-dominating sorting, and crowding distance selection~\cite{deb2002fast}. \RR{These strategies maintain the best possible approximation of the Pareto front with a limited size, thus avoiding memory and time overhead needed when incorporating, \eg{} the external archive concept \cite{Deb_MultiOOusingEA}.}   

\subsection{Update Strategy of Weighting Vectors}
\label{sec:newWeights}

A set of weighting vectors~$\M{W}_t$ is generated in every iteration~$t$. \RR{The goal is to find new weighting vectors that improve the quality of the Pareto front approximation. Our approach is inspired by~\cite{junqueira2022multi}, where the weights are updated based on the number of solutions in the local neighborhoods of the given weights. With the help of a powerful local algorithm that quickly identifies local extrema, we can focus on placing new weights only in unexplored areas of the Pareto front.}

The update strategy of the weighting vectors is summarized as pseudo-codes in Appendix~\ref{sec:algoWeightsUpdate}, see Algorithms~\ref{alg:locNeighborhoods} and~\ref{alg:finalWeights}. The first algorithm describes the calculation of the so-called local neighborhoods of weighting vectors and solutions from the previous iteration~$t-1$. The second algorithm presents a strategy to find a new set of weights used in the current iteration~$t$. Together, they serve as a tool to assign new areas where to search for the locally optimal solutions to enhance the quality of the found Pareto front approximation. 

\subsection{Assignment of Weights to Solutions}
\label{sec:asignWeights}

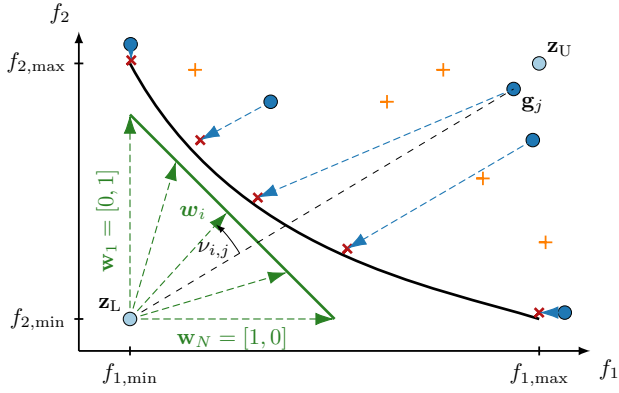
\begin{figure}
    \centering
    \scalebox{0.85}{



\begin{tikzpicture}
      \tikzstyle{weightsSty} = [-{Latex[length=3mm,width=2mm]}, PairedD, line width=0.5pt, densely dashed, cap=round, join=round]
      
      \tikzstyle{moveSty} = [-{Latex[length=3mm,width=2mm]}, >=stealth, scale=1, PairedB, line width=0.5pt, densely dashed, cap=round, join=round]
      
      \begin{customlegend}[
         legend columns=2,
         legend entries={ 
            weighting vector,
            previous generation,
            movement by loc. alg.,
            proposed by NSGA-II,
            utopian / nadir solution,
            locally optimal,
         },
         legend style={at={(3.7,7.0)}, anchor = north, column sep=10pt, font=\footnotesize}] 
         \addlegendimage{stealth-stealth,weightsSty}
         \addlegendimage{only marks,mark=+, color=PairedH, thick}
         \addlegendimage{stealth-stealth,moveSty}
         \addlegendimage{black,fill=PairedB, only marks,mark=*}
         \addlegendimage{black,fill=PairedA, only marks, mark=*}
         \addlegendimage{only marks,mark=x, color=PairedF, very thick}
      \end{customlegend}      
      
      \coordinate (O) at (0,0);
      \draw[thick,-latex] (O)  -- (8, 0) node[anchor=north west] {$f_1$};
      \draw[thick] (0.8 cm,2pt) -- (0.8 cm,-2pt) node[anchor=north] {$f_{1, \min}$};
      \draw[thick] (7.2 cm,2pt) -- (7.2 cm,-2pt) node[anchor=north] {$f_{1, \max}$};
      \draw[thick] (2pt, 0.5 cm) -- (-2pt, 0.5 cm) node[anchor=east] {$f_{2, \min}$};
      \draw[thick] (2pt, 4.5 cm) -- (-2pt, 4.5 cm) node[anchor=east] {$f_{2, \max}$};
      \draw[thick,-latex] (0,0) -- (0, 5.0) node[anchor=south east] {$f_2$};
      
      \coordinate (utop) at (0.8,0.5);
      \coordinate (nadir) at (7.2,4.5);
      \draw[very thick] (0.8,4.5) .. controls (2.4, 1.5) and (5.6,1.0) .. (7.2, 0.5);
      
      \coordinate (w1) at (0.8, 3.7);
      \coordinate (w2) at (1.52, 2.97);
      \coordinate (wMid) at (2.32, 2.17);
      \coordinate (w3) at (3.26, 1.23);
      \coordinate (wN) at (4.0, 0.5);
      
      \draw[weightsSty] (utop)  -- (wN);
      \draw[weightsSty] (utop)  -- (w1);
      \draw[weightsSty] (utop)  -- (wMid) node[left, xshift=-2mm] {$\boldsymbol{w}_{i}$};;
      \draw[weightsSty] (utop)  -- (w2);
      \draw[weightsSty] (utop)  -- (w3);
      \draw[very thick, PairedD] (0.8,3.7) -- (4.0, 0.5); 	
      
      \coordinate (g1) at (0.81,4.8);
      \filldraw[fill=PairedB] (g1) circle (3pt);
 
      \coordinate (g2) at (3.0,3.9);
      \filldraw[fill=PairedB] (g2) circle (3pt);
      \coordinate (gJ) at (6.8,4.1);
      \filldraw[fill=PairedB] (gJ) circle (3pt) node[below right] {$\M{g}_{j}$};
      \coordinate (g3) at (7.1,3.3);
      \filldraw[fill=PairedB] (g3) circle (3pt);
      \coordinate (gN) at (7.6,0.6);
      \filldraw[fill=PairedB] (gN) circle (3pt);
      
      \coordinate (n1) at (0.82,4.55);
      \draw[PairedF, very thick] (n1) -- ++(-2pt, -2pt) -- ++(4pt, 4pt) ++(-4pt, 0) -- ++(4pt, -4pt);
      \coordinate (n2) at (1.9,3.3);
      \draw[PairedF, very thick] (n2) -- ++(-2pt, -2pt) -- ++(4pt, 4pt) ++(-4pt, 0) -- ++(4pt, -4pt);
      \coordinate (nJ) at (2.8,2.4);
      \draw[PairedF, very thick] (nJ) -- ++(-2pt, -2pt) -- ++(4pt, 4pt) ++(-4pt, 0) -- ++(4pt, -4pt);
      \coordinate (n3) at (4.2,1.6);
      \draw[PairedF, very thick] (n3) -- ++(-2pt, -2pt) -- ++(4pt, 4pt) ++(-4pt, 0) -- ++(4pt, -4pt);
      \coordinate (nN) at (7.2,0.6);
      \draw[PairedF, very thick] (nN) -- ++(-2pt, -2pt) -- ++(4pt, 4pt) ++(-4pt, 0) -- ++(4pt, -4pt);
      
      \draw[moveSty] (g1)  -- (n1);
      \draw[moveSty] (g2)  -- (n2);
      \draw[moveSty] (gJ)  -- (nJ);
      \draw[moveSty] (g3)  -- (n3);
      \draw[moveSty] (gN)  -- (nN);
      
      \coordinate (p1) at (1.82,4.4);
      \draw[PairedH, thick] (p1) -- ++(-3pt, 0) -- ++(6pt, 0) ++(-3pt, 3pt) -- ++(0, -6pt);
      \coordinate (p2) at (4.82,3.9);
      \draw[PairedH, thick] (p2) -- ++(-3pt, 0) -- ++(6pt, 0) ++(-3pt, 3pt) -- ++(0, -6pt);
      \coordinate (pMid) at (5.7,4.4);
      \draw[PairedH, thick] (pMid) -- ++(-3pt, 0) -- ++(6pt, 0) ++(-3pt, 3pt) -- ++(0, -6pt);
      \coordinate (p3) at (6.32,2.7);
      \draw[PairedH, thick] (p3) -- ++(-3pt, 0) -- ++(6pt, 0) ++(-3pt, 3pt) -- ++(0, -6pt);
      \coordinate (pN) at (7.3,1.7);
      \draw[PairedH, thick] (pN) -- ++(-3pt, 0) -- ++(6pt, 0) ++(-3pt, 3pt) -- ++(0, -6pt);
      
      \draw[thin, dashed, black] (utop) -- (gJ); 	
      \draw
      pic [draw=black, -latex, "$\nu_{i,j}$", angle eccentricity=0.85, angle radius=2.0cm] {angle = gJ--utop--wMid};
      \filldraw[fill=PairedA] (utop) circle (3pt) node[above left] {$\M{z}_{\mathrm{L}}$};
      \filldraw[fill=PairedA] (nadir) circle (3pt) node[above right] {$\M{z}_{\mathrm{U}}$};
      \path let \p1 = (wN), \p2 = (utop) in coordinate (b) at (\x1/2+\x2/2 ,\y1);
      \node[below, PairedD] at (b) {$\M{w}_N = \left[1,0\right]$};
      \path let \p1 = (w1), \p2 = (utop) in coordinate (C) at (\x1-10 ,25 +\y1/2+\y2/2);
      \node[left, PairedD, rotate=90] at (C) {$\M{w}_1 = \left[0,1\right]$};

   \end{tikzpicture}
    }
    \caption{\RR{Association of weighting vectors~$\weight_i$ to solutions~$\gene_j \subset \mathcal{O}_t$ (dark blue circles) proposed by NSGA-II based on the previous generation~$\mathcal{G}_{t-1}$ (orange plus markers). Weighting vector~$\V{w}_i$ is associated with solution~$\V{g}_j$ based on similarity angle~$\nu_{i,j}$. The local algorithm then leads the proposed solutions to locally optimal solutions (red crosses) near the true Pareto front (black bold curve). The weighting vectors are convex, \ie{}, they form an~$M$-dimensional simplex (green bold curve).}} 
    \label{fig:WeightAssoc}
\end{figure}

At every iteration, a set of offspring shapes~$\mathcal{O}_t$ is created using the crossover \RR{(combination of two randomly selected shapes $\gene$)} and mutation \RR{(random alternation of one element of $\gene$)} operations (see step~7 in Fig.~\ref{fig:flowChart}), combining locally optimal shapes. To perform the local searches as effectively as possible, we apply a simple strategy to assign weights to individual shapes. A detailed description of the step-by-step algorithm can be found in Appendix~\ref{sec:algoWeights2Sol}. The main idea is visualized in Fig.~\ref{fig:WeightAssoc} showing the candidate solutions in the objective space. The orange plus markers denote the shapes from the previous generation~$\mathcal{G}_{t-1}$ that have been used to create a new set of initial shapes~$\mathcal{O}_{t}$ denoted by dark blue circles. \RR{Among the available shapes, the \Quot{utopian} solution~$\M{z}_{\mathrm{L}}$ and the \Quot{nadir} solution~$\M{z}_{\mathrm{U}}$ are identified~\cite{Deb_MultiOOusingEA}. The utopian solution combines the best possible values for all objective functions, while the nadir solution combines the worst. These hypothetical solutions are used for normalization and define the unit simplex for the weighting vectors, as shown by the bold green curve in Fig.~\ref{fig:WeightAssoc}. The weighting vectors are assigned to individual shapes based on similarity angle~$\nu$ in~\eqref{eq:thetaAngle}. After that, the local algorithm adjusts the initial shapes, moving them closer to the Pareto front in the objective space (indicated by the red cross markers in Fig.~\ref{fig:WeightAssoc}) along the direction defined by the assigned weight.} 

With respect to its extremely high efficiency, the local algorithm is the most computationally intensive step of the optimization process. During its execution, many new shapes are explored. \RR{To ensure efficient optimization, the weighting vector assigned to a shape should not point toward one end of the Pareto front if the shape is located near the other end. Such a mismatch would require the local algorithm to significantly alter the antenna's design, making the process inefficient. At the same time, it is important to prevent agents from repeatedly exploring the same regions of the Pareto front in consecutive iterations. By combining Algorithms~\ref{alg:finalWeights} and~\ref{alg:weightAssign}, we achieve a balance between refining known solutions (exploitation) and discovering new ones (exploration). Solutions with the smallest angular distance to a given weight are used to fine-tune the corresponding region of the Pareto front. In contrast, solutions with larger angular distances are directed toward unexplored areas, promoting diversity in the search.}

\section{Examples}
\label{sec:examples}

Several challenging optimization problems are presented in this section to demonstrate the proposed algorithm's superb performance. To assess the quality of the proposed algorithm, three strategies are compared.
\begin{description}[\IEEEsetlabelwidth{MOMA-AW}]
    \item[MOMA-AW:] The algorithm proposed in this work. It is based on adaptive weighting and the memetic optimizer combining the gradient step and NSGA-II.
    \item[SOGA-FW:] An iteratively restarted memetic optimizer combining the gradient step and GA with \textit{a priori} chosen and swept weights.
    \item[NSGA-II:] A state-of-the-art MO algorithm without the use of the local step.
\end{description}

The optimization simulations are performed within the parameters of GA as summarized in Table~\ref{Tab:FOPSset}. The values are set so that the exploration of NSGA-II is favored due to the exceptional convergence speed of the underlying local algorithm. Unless otherwise specified, the underlying local algorithm always runs until it reaches a local minimum. All tests are performed on an ALUCARD server with an AMD Ryzen Threadripper 3990X 64-Core Processor with 220~GB RAM. All optimization tasks are parallelized, so each CPU thread performs one local optimization at a time.

The performance of the algorithms is assessed using standard multi-objective metrics, specifically generational distance (GD) and hypervolume (HV) \cite{Deb_MultiOOusingEA}. Generational distance measures the mean distance of the non-dominated solutions found \RR{by the assessed optimizer to the closest point on a true Pareto front. Hypervolume measures the amount of the feasible objective space dominated by the non-dominated solutions found.} Ideally, the optimal solution would have a~$\mathrm{GD}$ approaching zero and the largest possible~$\mathrm{HV}$ value. The equations for calculating these metrics are summarized in Appendix~\ref{sec:metricsMO}. Each study is performed 30~times to suppress the stochastic effects, reporting the mean, best, and worst results.

\begin{table}[t]
\caption{Parameter settings of MOMA-AW, SOGA-FW, and NSGA-II algorithms used in this paper. For a detailed explanation of the parameters, see~\cite{marek2020fops}.}
\centering
\begin{tabular}{>{\centering\arraybackslash}p{4.5cm}>{\centering\arraybackslash}p{2.5cm}}
\multicolumn{1}{c}{parameter}                    & \multicolumn{1}{c}{value $\mathrm{(-)}$} \\ \toprule 
number of agents, $N_{\mathrm{A}}$               & 64  \\ 
number of iterations, $N_{\mathrm{I}}$           & 40 \\
probability of mutation, $p_{\mathrm{M}}$        & 1  \\
probability of crossover, $p_{\mathrm{C}}$       & 0.9  \\
number of crossover positions, $N_{\mathrm{CP}}$ & 1  \\ \bottomrule
\end{tabular}%
\label{Tab:FOPSset}
\end{table}

\subsection{Q Factor and Electrical Size}
\label{sec:ExA}

The first example deals with the trade-off between the Q~factor, which is a good indicator of the fractional bandwidth of an electrically small antenna~\cite{YaghjianBest_ImpedanceBandwidthAndQOfAntennas} and electrical size. As widely reported in the literature, the trade-off between the Q~factor and electrical size is approximately cubic for electrically small sizes~\cite{Chu_PhysicalLimitationsOfOmniDirectAntennas, Capek_etal_2019_OptimalPlanarElectricDipoleAntennas}, \ie{}, $Q \propto 1/(ka)^3$, where~$k$ is the vacuum wave number, and~$a$ is the radius of the sphere \RR{that fully circumscribes} an antenna. Such extreme scaling is an excellent initial test of the proposed MO algorithm with inhomogeneous and adaptively updated optimization weights.

\begin{figure}[t]
    \centering
    \includegraphics[width=\columnwidth]{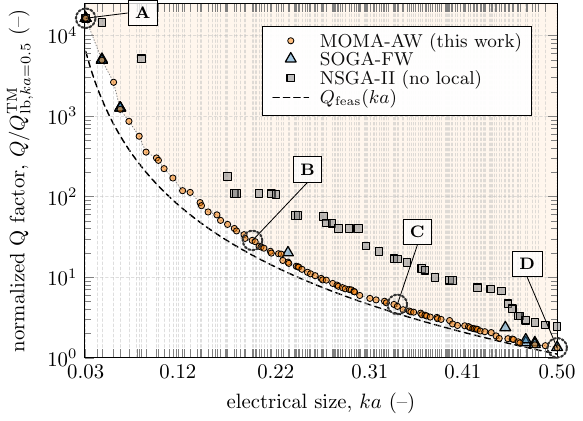}
    \caption{Multi-objective topology optimization of the Q~factor and the electrical size~$ka$. The design region is a rectangular plate of the side aspect ratio~2:1 with~$16 \times 8$ pixels ($744$~basis functions). The structure is fed by a discrete feeder in the middle of the longer edge, see Fig.~\ref{fig:ExA2}. Different sets of markers represent different optimization methods. The method proposed in this paper (MOMA-AW) has the non-dominated solutions denoted by circles. The black dashed line shows the asymptote~$1/(ka)^3$ normalized so that it is equal to the value~$Q_{\T{min}, ka=0.5}/Q_\T{lb}^\T{TM}=1.11$ at~$ka=0.5$, where~$Q_{\T{min},ka=0.5}$ is the minimum Q factor value found by topology optimization. All traces are normalized to the Q factor lower bound evaluated for~$ka = 0.5$ ($Q_\T{lb}^\T{TM} = 42.2$) with only TM modes considered~\cite{Capek_etal_2019_OptimalPlanarElectricDipoleAntennas}. Four diverse solutions are tagged by a letter, and the associated candidates are shown in Fig.~\ref{fig:ExA2}. The best run out of 30 runs (in terms of maximum HV) is shown for all three optimization methods.}
    \label{fig:ExA1}
\end{figure}

\begin{figure*}[t]
    \centering
    \includegraphics[width=\textwidth]{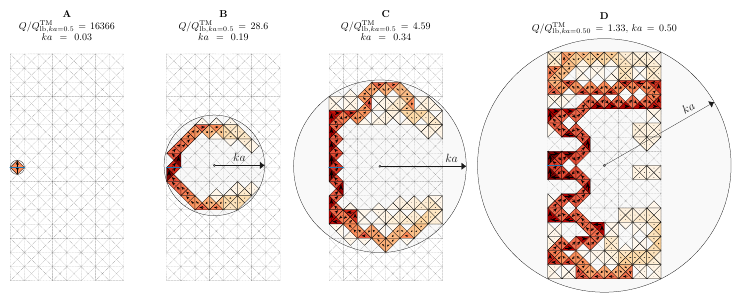}
    \caption{Four Pareto-optimal candidates from Fig.~\ref{fig:ExA1}, labeled~A-D, are shown from left to right. Their normalized Q~factor performance and electrical size are shown on the top. The electrical size of the smallest circumscribing sphere is highlighted by the black circle.}
    \label{fig:ExA2}
\end{figure*}

The optimization is performed over a composite objective function
\begin{equation}
    \widetilde{f}_{\T{A}} (\srcRegion_n) = w_1 \dfrac{Q_n}{Q_\T{lb}^\T{TM}} + w_2 \dfrac{a_n}{a_0}, \quad \sum_{i} w_i = 1,
    \label{eq:ex1}
\end{equation}
where $Q_n$ is the Q~factor of an antenna represented by shape~$\srcRegion_n$, evaluated via~\cite[(1)]{2021_capeketal_TSGAmemetics_Part2}, and normalized to the lower bound~$Q_\T{lb}^\T{TM}$, \cite{CapekGustafssonSchab_MinimizationOfAntennaQualityFactor, Capek_etal_2019_OptimalPlanarElectricDipoleAntennas}, \ie{}, $Q_n/Q_\T{lb}^\T{TM} \in [1, \infty)$. The electrical size of antenna~$\srcRegion_n$ is given by radius~$a_n$ of the smallest sphere fully circumscribing the antenna, normalized to the radius of the design region~$a_0$, \ie{}, $a_n/a_0 \in [0,1]$. Notice that actual size~$a_n$ is evaluated using distance matrix~$\M{D}$ which is precalculated at the beginning of the optimization, see Appendix~\ref{sec:distMat}. Since the density of the mesh grid is fixed for the entire optimization, a sufficiently fine grid has to be used to ensure that all physical quantities are evaluated correctly. 

The proposed approach, with electrical size~$ka$ being one of the optimized parameters, \cf{}, \eqref{eq:ex1}, makes it possible to run the entire optimization only once and obtain optimal shapes in the Q factor for all electrical sizes~$ka$ at once. Changing the spatial dimension ($a$) allows us to fix the frequency ($k$) and calculate all required method-of-moment matrices only once~\cite{2021_capeketal_TSGAmemetics_Part1, 2021_capeketal_TSGAmemetics_Part2} before the optimization.

The optimization setup is as follows. The design region with an aspect ratio of 2:1 is segmented into $16 \times 8$ squares, and each square is triangularized into 4 identical triangles. The material is the perfect electric conductor. The total number of basis functions is~744. The discrete feeder is connected to the middle of the long side, close to the boundary of the design region, \ie{}, there are 743 discrete optimization variables. The discretizations of $12 \times 6$ and $20 \times 10$ (414 and 1170 basis functions, respectively) were also considered to study convergence properties.

\begin{figure}[t]
    \centering

    \includegraphics[width=\columnwidth]{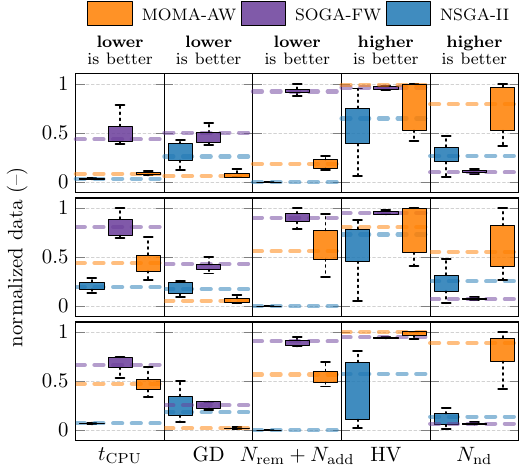}    
    \caption{\RR{Boxplots of the normalized metrics for Example~A and grid size $12 \times 6$ (top pane), $16 \times 8$ (middle), and $20 \times 10$ (bottom). The solid vertical line separates metrics that should be minimized: execution time $t_{\T{CPU}}$, generational distance GD, number of perturbations $N_{\T{rem}} + N_{\T{add}}$; and maximized: hypervolume HV, and number of non-dominated solutions found $N_{\T{nd}}$.}} 
    \label{fig:boxplotA}
\end{figure}

MOMA-AW makes it possible to find the entire Pareto front depicted by the circular markers in Fig.~\ref{fig:ExA1} at once. On the contrary, the conventional way to investigate the trade-offs is to set the weights and run SO repeatedly sequentially. This is depicted as the SOGA-FW technique with the triangle markers in Fig.~\ref{fig:ExA1}. It can be seen that performance is considerably worse in terms of quality (how regularly the front is sampled and how the solutions are dominated). This observation is confirmed \RR{by Fig.~\ref{fig:boxplotA} comparing GD and HV parameters and overall computational time in the form of standard boxplots for all the compared algorithms. For the statistical data, please refer to Table~I in the supplementary file ~\cite{KadlecCapek2024}.} Another possible approach is to take the NSGA-II MO algorithm and apply it to the problem~\eqref{eq:ex1} without any local (rank-1) updates. This results in the square markers in Fig.~\ref{fig:ExA1}. While the number of found candidates is satisfactory for larger~$ka$, the quality of the solutions is poor (Q factor is approximately one order worse than that found with MOMA-AW).

Thanks to the rich history of Q factor minimization~\cite{2018_Schab_Wsto}, this example allows us to compare the results with the fundamental bounds~\cite{CapekGustafssonSchab_MinimizationOfAntennaQualityFactor} determined for every~$ka$. When only TM modes are considered, as is usually done for electrically small scenarios with one feeder~\cite[Fig.~6]{CapekGustafssonSchab_MinimizationOfAntennaQualityFactor}, we get the $Q_\T{lb}^\T{TM} (ka)$ bound. To approximately demarcate the feasible region, the value of the bound is renormalized with the performance of the best antenna found at $ka = 0.5$ with the Q factor value~$Q_{\T{min},ka=0.5}$, \ie{},
\begin{equation}
    Q_\T{feas} (ka) = \dfrac{Q_{\T{min}, ka=0.5}}{Q_{\T{lb}, ka=0.5}^\T{TM}} Q_\T{lb}^\T{TM} (ka)
\end{equation}
and depicted in Fig.~\ref{fig:ExA1} as the black dashed line.

The fact that, in this case, the Q~factor bound is tight and is known~\cite{Capeketal_InversionFreeEvaluationOfNearestNeighborsInMoM, 2021_capeketal_TSGAmemetics_Part2}, allows us to populate the entire Pareto front within one optimization run. This contrasts with all state-of-the-art algorithms where optimal antennas are extracted separately one by one for all frequencies~\cite{CismasuGustafsson_FBWbySimpleFreuqSimulation}.

It is seen in Fig.~\ref{fig:ExA1} that the solutions found by MOMA-AW follow the boundary of feasible region~$Q_\T{feas} (ka)$ well. Still, there is an increasing gap between the bound and the realized antennas for lower $ka$ (the orange circle marks and the black dashed line) due to the fixed discretization grid. The smaller electrical size requires higher spatial resolution to obtain a good solution~\cite{Capek_etal_2019_OptimalPlanarElectricDipoleAntennas}, but, in this example, the opposite is the case as the smaller electrical size leads to fewer degrees of freedom available. 

In terms of MO metrics \RR{compared in Fig.~\ref{fig:boxplotA}}, the proposed MOMA-AW algorithm outperforms the other two algorithms in~$\T{GD}$. MOMA-AW achieves average~$\T{GD}$ values ($\mathrm{GD}_{\mathrm{avg}}$) over 30 runs that are nearly one order of magnitude better than SOGA-FW, and two orders better than NSGA-II. However, the comparison using the~$\T{HV}$ metric is not as straightforward. The average~$\T{HV}$ value for SOGA-FW is slightly better than that of MOMA-AW across all grid sizes. This is due to the highly non-linear nature of the Pareto front where the solutions in the top-left section (high Q factor, low $ka$) have the most significant influence on the~$\T{HV}$ value but are, in fact, very poor antennas. While SOGA-FW consistently finds a few solutions in this critical region, MOMA-AW sometimes fails to find solutions with electrical size~$ka \le 0.15$ (except for the extreme solution labeled A in Fig.~\ref{fig:ExA1}). Despite this, the overall quality of the Pareto front found by MOMA-AW is much better than that found by SOGA-FW. For example, considering the best solutions from 30 independent runs shown in Fig.~\ref{fig:ExA1}, all solutions found by SOGA-FW (denoted by triangles in Fig.~\ref{fig:ExA1}) are dominated, \ie{}, they are strictly worse in at least one of the metrics compared to solutions found by MOMA-AW. On the contrary, no solution found by MOMA-AW is dominated by a solution proposed by SOGA-FW.    

Four antenna samples of differing $ka$~size are seen in Fig.~\ref{fig:ExA2} which shows the evolution of optimal shape with regard to electrical size. The antennas, labeled A-D, correspond to the positions at the Pareto front in Fig.~\ref{fig:ExA1}. The smallest antenna,~A, is, in fact, the isolated feeder with two adjacent triangles. It is the smallest \Quot{antenna} possible with the given mesh grid. Therefore, its performance delimits the Pareto front from the left, see Fig.~\ref{fig:ExA1}. More realistic cases are generated for higher~$ka$, namely antennas~B and~C in Fig.~\ref{fig:ExA2}. The shapes are bent dipoles, slowly evolving into meanderline, as represented by sample D. The algorithm finds almost 100 Pareto-optimal candidates offering many possible shapes for the final design, see the supplementary material~\cite{KadlecCapek2024} for all non-dominated solutions. It should be noted that not all these antenna shapes are ready for manufacturing. Final adjustments must be done by antenna engineers.

\begin{figure}[t]
    \centering
    \includegraphics[width=8.9cm]{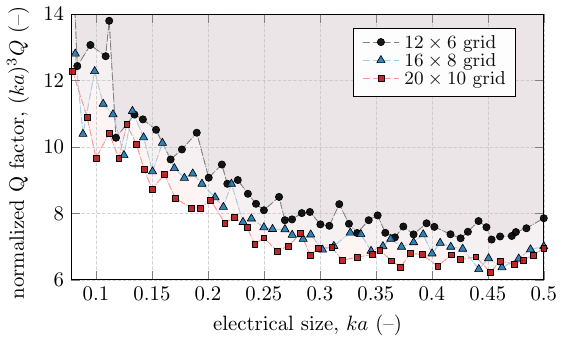}
    \caption{A comparison of minimum Q factor values achieved at different electrical sizes~$ka$ and for different mesh grid densities. Each trace was calculated within a single run of MOMA-AW. The best Pareto front is always taken out of 30 independent runs based on the~$\T{HV}_\T{max}$ parameter. The values of the Q factor are normalized by multiplying~$(ka)^3$ which is the scaling of the Q factor,~\cite{Capek_etal_2019_OptimalPlanarElectricDipoleAntennas}. The feasible areas are shaded and correspond to the color of the markers.}
    \label{fig:ExA3}
\end{figure}

\begin{figure}[t]
    \centering
     \includegraphics[width=\columnwidth]{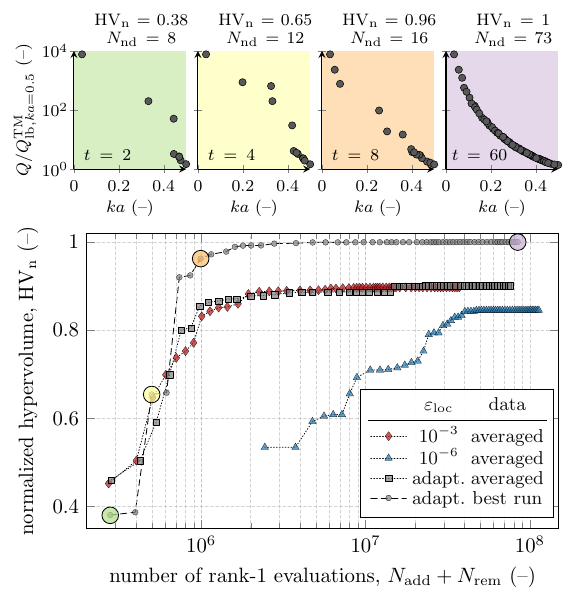}
    \caption{Study of hypervolume (HV) maximization as a function of required \mbox{rank-1} perturbations~$N_\T{add} + N_\T{rem}$ (number of additions and removals). A mesh grid of $16 \times 8$~pixels and electrical size~$ka = 0.5$ was used. Hypervolume is normalized to the maximum value achieved in the study. Three different termination criteria for the local algorithm are studied and averaged by running the same study 30 times. The best run achieved is separately depicted as the dashed line with circular markers, and the corresponding Pareto fronts existing at that stage of the multi-objective optimization are depicted in the top panes, where~$t$ denotes the global iteration, and~$N_\T{nd}$ denotes the number of non-dominated solutions.}
    \label{fig:ExA5}
\end{figure}

To check the convergence of the method, three mesh grids of various granularity were compared, see Fig.~\ref{fig:ExA3}. As expected, higher spatial resolution leads to better performance evidenced by the red curve, as compared with the black curve in Fig.~\ref{fig:ExA3}. This is, however, at the cost of additional computational time, \cf{}, \RR{Fig.~\ref{fig:boxplotA}}. The normalized Q for lower~$ka$ reaches value~6, which is close to heuristically designed meanderlines~\cite{Capek_etal_2019_OptimalPlanarElectricDipoleAntennas}. For lower~$ka$, performance deteriorates but remains generally better with the finer grid.

It is seen from \RR{Fig.~\ref{fig:boxplotA}} that the most expensive part of the MOMA-AW algorithm is the local gradient-based algorithm, \cf{}, computational time $t_\T{cpu}$ of NSGA-II, and MOMA-AW. On the other hand, thanks to this step, the global algorithm (GA) moves only in the sub-space of local minima. The process of acquiring sufficient hypervolume~$\mathrm{HV}$ is studied in Fig.~\ref{fig:ExA5} and depends on the termination criterion of local algorithm~$\varepsilon_\T{loc}$. Parameter~$\varepsilon_\T{loc}$ controls the relative improvement of the objective function under which the local optimization is terminated and the solution found is returned~\cite{2021_capeketal_TSGAmemetics_Part1}. It is seen that the best strategy is to taper the relative precision so that it is rather low at the beginning (say, $\varepsilon_\T{loc}= 10^{-3}$ for $t < 10$) then slowly decrease it~(reaching $\varepsilon_\T{loc}= 10^{-6}$ for $t = 30$). The local algorithm thereby spends less effort fine-tuning the local optima at the beginning and gradually creates more refined candidates that are closer to the true Pareto front. Also, we see in the top panes of Fig.~\ref{fig:ExA5} that the Pareto front is approximately covered at a relatively early stage of the optimization (iteration~$t = 8$), but it takes many iterations and local updates to cover the Pareto front uniformly.

\subsection{Q Factor, Impedance Matching, and Shape Regularity}
\label{sec:ExB}

This example deals with the trade-off between the Q factor (bandwidth), impedance matching, and the regularity of antenna shape. As shown in the previous example, the Q~factor is severely constrained by electrical size, therefore, it is one of the main limitations for electrically small antennas. Good impedance matching is often advantageous in these cases, \cite{Best_ElectricallySmallResonantPlanarAntennas, Fujimoto_Morishita_ModernSmallAntennas}. Another problem is the questionable practicability of antennas found with heuristic algorithms~\cite{RahmatSamii_Kovitz_Rajagopalan-NatureInspiredOptimizationTechniques}. They are often highly irregular and hard to manufacture~\cite{CismasuGustafsson_FBWbySimpleFreuqSimulation, YangAdams_SystematicShapeOptimizationOfSymmetricMIMOAntennasUsingCM, 2021_capeketal_TSGAmemetics_Part2}. To resolve this issue, a set of shape regularity parameters was proposed in~\cite{Capek_etal_RegularityConstraints_EuCAP2021}. These parameters are inexpensive to evaluate and fully compatible with the MoM paradigm and rank-1 updates produced by the local algorithm.

The composite objective function, in this case, reads
\begin{equation}
    \label{eq:exB}
    \widetilde{f}_{\T{B}} (\srcRegion_n) = w_1 \dfrac{Q_n}{Q_\T{lb}^\T{TM}} + w_2 |\varGamma_n|^2 + w_3 R_n, \quad \sum_i w_i = 1,
\end{equation}
where $\varGamma_n$ is the reflection coefficient of antenna~$\srcRegion_n$ defined as
\begin{equation}
    \varGamma_n = \dfrac{Z_{\T{in},n} - Z_0}{Z_{\T{in},n} + Z_0},
\end{equation}
with $Z_0$ being the reference impedance, here set to $Z_0 = 20\,\Omega$,
\begin{equation}
    Z_{\T{in},n} = \dfrac{V_{\T{in}}}{I_{\T{in},n}}
\end{equation}
being the input impedance of the antenna, see \cite[(4)]{2021_capeketal_TSGAmemetics_Part2} for details, and~$R_n$ controls the regularity of a shape as
\begin{equation}
    \label{eq:exBreg}
    R_n = 0.15 \dfrac{A_n}{A_0} + 0.30 h_n,
\end{equation}
where $A_n/A_0$ is the relative area of the design region spanned by metallization~$A_n$ as compared to the area of design region~$A_0$, and $h_n$ describes the smoothness of the shape~\cite[(13)]{Capek_etal_RegularityConstraints_EuCAP2021}.

\RR{Regularity parameter~$R \in (0, 0.45)$ in~\eqref{eq:exBreg} constraints occurrence of artifacts associated with discrete topology optimization~\cite{Capek_etal_RegularityConstraints_EuCAP2021} such as point connections~\cite{Jayasinghe2015}, isolated islands of material~\cite[Figs. 5 and 8]{CismasuGustafsson_FBWbySimpleFreuqSimulation}, or infinitely thin slots~\cite{Capeketal_ShapeSynthesisBasedOnTopologySensitivity}. These phenomena require tedious post-processing, which also makes the optimized structures suboptimal. Depending on the antenna designer's preference, the regularity constraint might be taken into account within MO optimization only partly (any value of $R < R_\T{user}$ is tolerated) or ignored, reducing the dimensionality of the Pareto front by one.}

The design region is a triangularized rectangular shape defined by~$16 \times 8$ pixels of electrical size~$ka =0.5$. The material is a perfect electric conductor. The single delta gap feeder with~$V_{\T{in}} = 1\,$V is connected as in the previous example.

\begin{figure*}[t]
    \centering
    \includegraphics[width=0.85\textwidth]{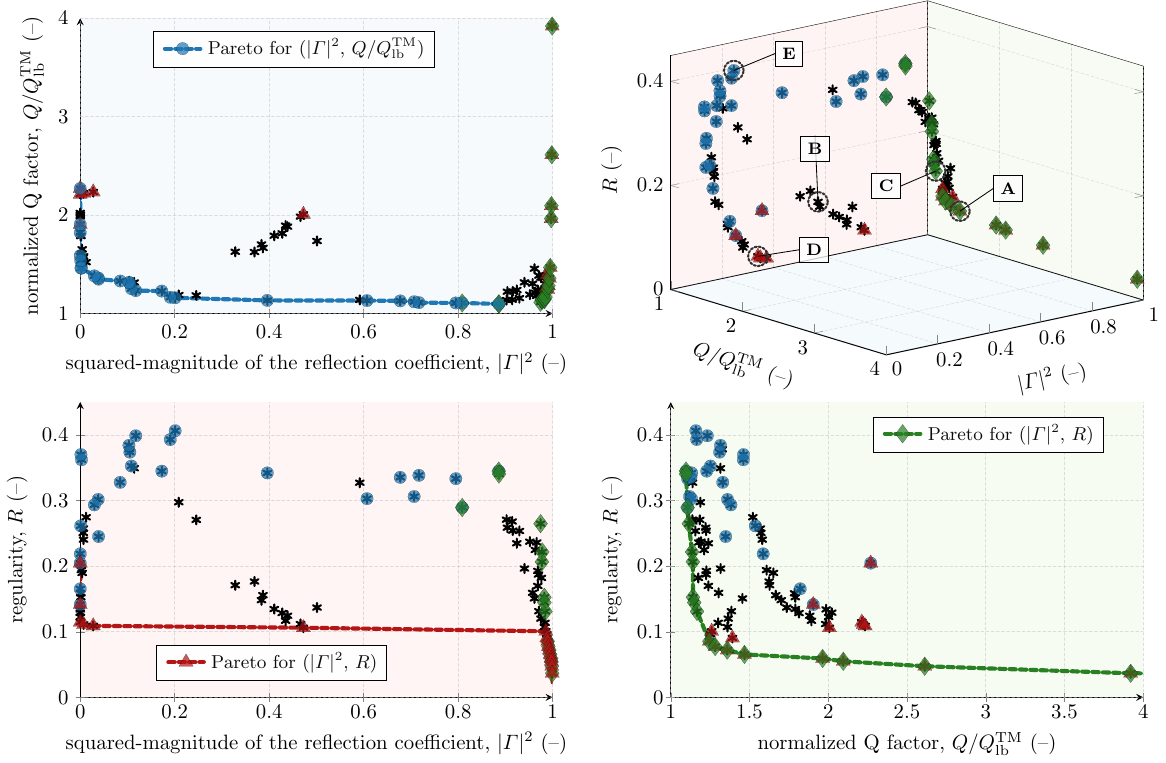}
    \caption{Non-dominated solutions for the multi-objective optimization problem of minimizing Q factor~$Q/Q_\T{lb}^\T{TM}$, reflection coefficient~$\varGamma$, and improving regularity~$R$. The values of the Q factor are normalized as in Fig.~\ref{fig:ExA1}. The reference impedance is $Z_0 = 20\,\Omega$. Three principal cuts show the pairs of optimized objectives. The cutting planes are assigned different colors. The non-dominated solutions for a given cutting plane are highlighted by the same color as the background filling and are connected by a dashed line. The top-right pane depicts the three-dimensional plot. Five diverse solutions are chosen from the top-right pane and depicted in Fig.~\ref{fig:ExB2}. The best run out of 30 runs (in terms of maximum HV) is shown.}
    \label{fig:ExB1}
\end{figure*}

The MO problem~\eqref{eq:exB} has three independent objectives and this poses a problem when presenting final results. The Pareto front is depicted in Fig.~\ref{fig:ExB1} both in terms of a 3D plot in the top-right pane and as two-dimensional cuts for all combinations of optimized parameters. All three parameters mutually conflict since the Pareto fronts associated with all optimized parameter pairs form a curve. The most challenging trade-off seems to be between matching and regularity, see the red dashed curve in the bottom-left pane of Fig.~\ref{fig:ExB1}, which indicates that good matching requires considerable irregularity of the antenna (\eg{}, meanderline with many meanders, \cite{Capek_etal_2019_OptimalPlanarElectricDipoleAntennas}). The non-dominated solutions jump abruptly from matched well to poorly matched, with only one exception around~$|\varGamma|^2 \approx 0.45$.

\begin{figure*}[t]
    \centering
    \includegraphics[width=0.9\textwidth]{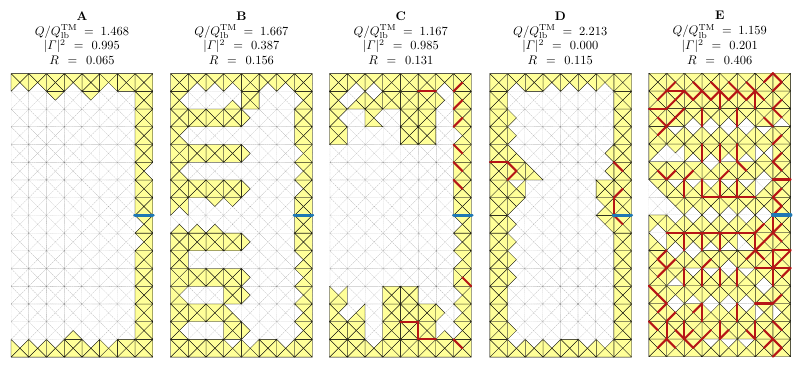}
    \caption{Antenna candidates chosen from the Pareto front of Example~B depicted in the top-right pane of Fig.~\ref{fig:ExB1}, labeled~A-E. All antenna samples are fed by the discrete feeder at the position highlighted by the thick blue edge (in the middle on the right). The optimized performance parameters are shown at the top. The red edges are removed and, in practice, should be replaced by a thin slot~\cite{Capeketal_ShapeSynthesisBasedOnTopologySensitivity}.}
    \label{fig:ExB2}
\end{figure*}

As in Section~\ref{sec:ExA}, several antenna candidates from the Pareto front in Fig.~\ref{fig:ExB1} are labeled, with their shape and performance depicted in Fig.~\ref{fig:ExB2}. The rest of the non-dominated solutions are presented in the supplementary material~\cite{KadlecCapek2024}. The excitation is highlighted by the blue edge, the material segments are the yellow triangles, the vacuum is represented by the white triangles, and the thick red edges are associated with optimized degrees of freedom replaced by a thin slot~\cite{Capeketal_ShapeSynthesisBasedOnTopologySensitivity}. Because of the trade-off, there is no free lunch for the antenna design -- the shapes have either a good Q factor, see shapes ~C and~E, but with poor matching as in~C, or poor regularity as in~E, or they have good matching, as in shape~D, but with a relatively high Q factor. Real antenna design is usually a compromise between conflicting parameters, as shown in Fig.~\ref{fig:ExB1}, and, as such, the MOMA-AW algorithm can effectively inform the designer about the trade-offs and possible shapes to initiate the design process.

\begin{figure}
    \centering

    \includegraphics[width=\columnwidth]{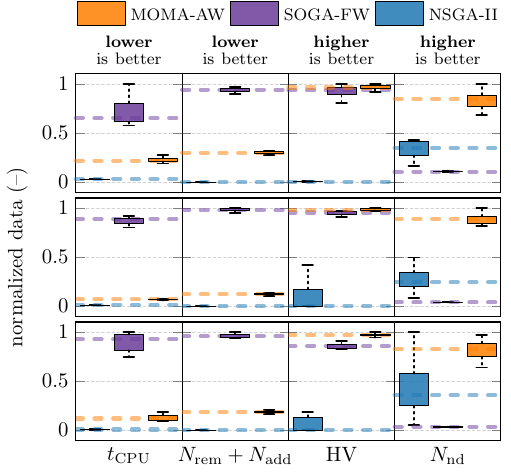}
    \caption{\RR{Boxplots of the normalized metrics for Example~B and grid size $12 \times 6$ (top pane), $16 \times 8$ (middle), and $20 \times 10$ (bottom). We aim to minimize the metrics on the left side of the solid black vertical line (computational time, and number of perturbations) while maximizing those on the right (hypervolume, and number of non-dominated solutions), ensuring a balanced optimization approach.} } 
    \label{fig:boxplotB}
\end{figure}

The performance of all three algorithms used in this paper is compared in \RR{Fig.~\ref{fig:boxplotB}, see the supplementary material~\cite[Table~II]{KadlecCapek2024} for tabulated data}. Notice that the generational distance is not depicted in this case since the ground truth, compared to the previous example, is unknown. Only five independent runs of SOGA-FW were executed due to its enormous time burden. The average time needed for one run of SOGA-FW was approximately $t_{\mathrm{cpu}} = 47.5 \,\mathrm{hours}$. Our algorithm MOMA-AW performs consistently better in all parameters. The only exception is the computational time comparing MOMA-AW to the NSGA-II. Notice, however, that the quality of the Pareto front found by NSGA-II is worse by one or two orders in the HV parameter. The difference in Pareto front quality between MOMA-AW and SOGA-FW increases with increasing grid size and the increasing number of DOF. For a grid with size $20 \times 10$, the worst result found by MOMA-AW ($\T{HV}_{\min} = \num{7.23e3}$) is still better than the best solution found by SOGA-FW ($\T{HV}_{\max} = \num{6.99e3}$). Moreover, the SOGA-FW performed, on average, more than a billion perturbations (almost an order of magnitude more than MOMA-AW) and still was able to find only 17 non-dominated solutions, while MOMA-AW was able to find, on average, more than 400~non-dominated solutions and cover the entire Pareto series much better. This trend is expected to continue with an increasing number of DOF.  

\subsection{Realized Gain, Electrical Length, and Shape Regularity}
\label{sec:ExC}

\begin{figure*}[t!]
    \centering
    \includegraphics[width=0.85\textwidth]{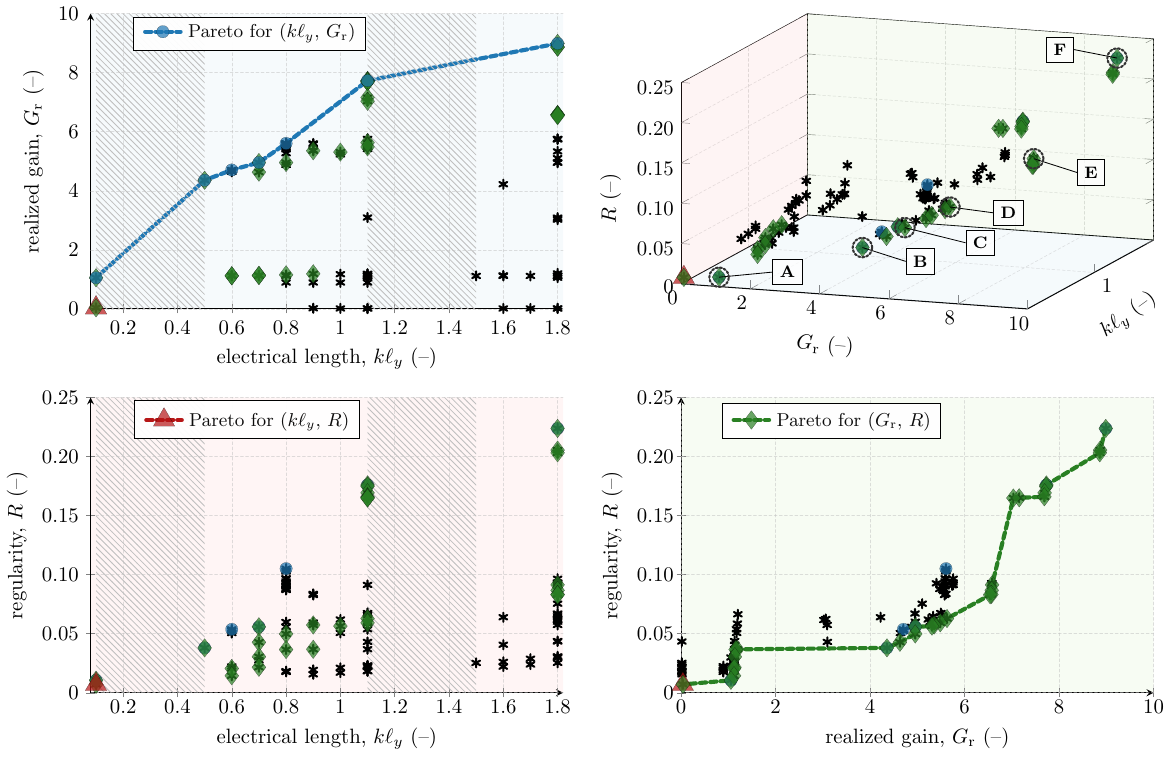}
    \caption{Non-dominated solutions for Example~C: minimizing the electrical length of \RR{an antenna}~$k L_y$, maximizing realized gain~$G_\T{r}$, and improving irregularity~$R$. The graphical style of the figure is the same as for Fig.~\ref{fig:ExB1}. Six solutions are chosen and depicted in Fig.~\ref{fig:ExC2}. They are denoted as A-F in the top-right pane. The best run out of 30 runs (in terms of maximum HV) is shown.}
    \label{fig:ExC1}
\end{figure*}

\begin{figure}[t!]
    \centering
    \includegraphics[]{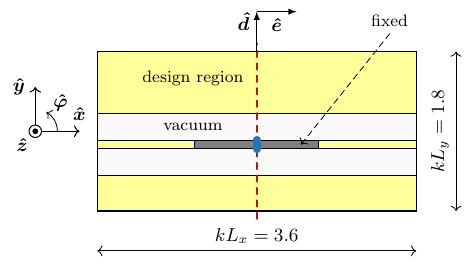}
    \caption{Sketch of the design region used for the optimization of~\eqref{eq:probC}. The red dashed line highlights the applied perfect electric conducting symmetry plane. The fixed region (not optimized) is filled with a gray color, and the excitation is represented by the blue oval marker.}
    \label{fig:ExC3}
\end{figure}

\begin{figure*}[t!]
    \centering
    \includegraphics[scale=1.4]{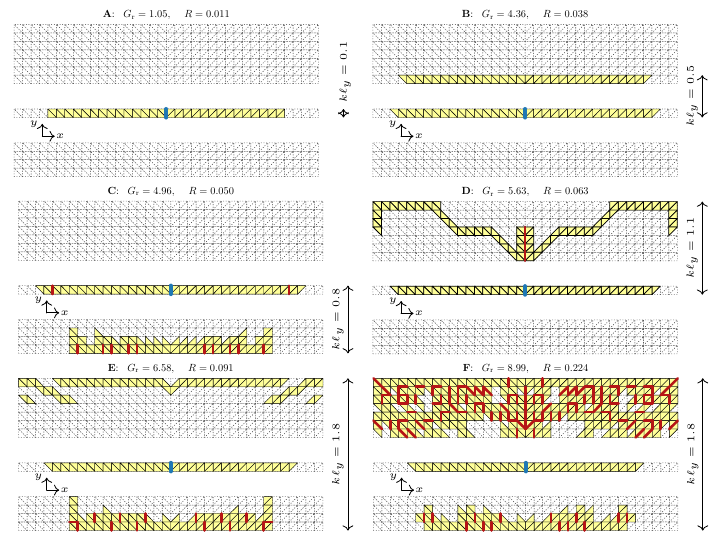}
    \caption{Antenna candidates picked from the Pareto front in the top-right pane of Fig.~\ref{fig:ExC1}. All antenna samples are fed by the discrete feeder at the position highlighted by the thick blue edge (in the middle of the dipole). The performance is shown on the top, and the electrical length is on the right. The central part of the dipole was fixed during the optimization. Its length and the area of the director(s) and reflector(s) are the subject of the topology optimization. The left-right symmetry was enforced to reduce the number of optimization variables. The red edges are removed and, in practice, should be replaced by a thin slot~\cite{Capeketal_ShapeSynthesisBasedOnTopologySensitivity}. The electrical size of one discretization pixel is $k\ell = 0.1$.}
    \label{fig:ExC2}
\end{figure*}

The \RR{next} example studies the maximal realized gain dependent on the electrical length of a \RR{high-gain antenna} and its regularity. Interest in \RR{supergain antennas} has continued in recent decades~\cite{UzsokySolymar_TheoryOfSuperDirectiveLinearArrays, AltshulerODonnellYaghjianBest_AMonopoleSuperdirectiveArray, Yaghjian_etal_RS2008, Debard2023}. Nevertheless, the majority of scientists have focused on uniform arrays \RR{and antennas} with predefined shapes or have utilized simple genetic algorithms.

To stimulate further research in this direction, we assume the following objective function
\begin{equation}
    \widetilde{f}_{\T{C}} (\srcRegion_n) = - w_1 G_{\T{r},n} + w_2 \dfrac{k \ell_{y,n}}{k L_y} + w_3 R_n,
    \label{eq:probC}
\end{equation}
where $G_\T{r}$ is the realized gain defined as
\begin{equation}
    \label{eq:refGain}
    G_{\T{r},n} = \eta_{\T{rad},n} \left( 1 - |\varGamma_n|^2 \right) D_n (\UV{d},\UV{e}),
\end{equation}
with~$\eta_{\T{rad},n}$ being radiation efficiency, $\varGamma_n$~being the reflection coefficient, and $D_n (\UV{d},\UV{e})$ being antenna directivity pointing towards direction~$\UV{d}$ with polarization~$\UV{e}$. The Pareto front is shown in Fig.~\ref{fig:ExC1} with the same layout as in Fig.~\ref{fig:ExB1}. The electrical size of an \RR{antenna} is taken into account as~$\ell_{y,n}/L_y$, where~$\ell_{y,n}$ is the length of an actual shape~$\srcRegion_n$ and~$L_y$ is the length of the design region. To improve manufacturability, regularity is once more assumed with parameter~$R_n$ defined according to~\eqref{eq:exBreg}.

The arrangement is as follows. The design domain lies in the xy plane of electrical width $k L_x = 3.6$ and electrical length~$k L_y = 1.8$ and is discretized into~$36 \times 18$ pixels and triangularized afterward. The region is separated into three areas, see Fig.~\ref{fig:ExC3}. The top and bottom areas can be freely modified. The middle area is partly fixed with the discrete feeder in the middle. The reference impedance is $Z_0 = 20\,\Omega$. Radiation is maximized along $\UV{d} = \UV{y}$ axis with polarization $\UV{e} = \UV{x}$. The material used for optimization is copper with conductivity~$\sigma = 5.96\cdot 10^7\,\T{Sm}^{-1}$. To lower the computational burden, the point group symmetry~\cite{SchabEtAl_EigenvalueCrossingAvoidanceInCM, Maseketal_ModalTrackingBasedOnGroupTheory}, with respect to the yz-plane, is applied, reducing the dimensions of the matrices and the number of optimization unknowns approximately by half (593~binary optimization variables). As compared to the previous examples, 128 GA agents were used in this example to boost the convergence.

Much of the insight can be extracted from the two-dimensional cuts depicted in Fig.~\ref{fig:ExC1}. As expected, there is a remarkable trade-off between the electrical length and realized gain, see the top-left pane. When the electrical length is restricted to $k\ell_y = 0.1$, \ie{}, to a sole dipole, the realized gain cannot exceed 1.64 (directivity of a dipole). Due to the structure of the design domain, see Fig.~\ref{fig:ExC3}, there are two gaps in $k\ell_y$, $k\ell_y \in (0.1, 0.5)$ and $k\ell_y \in (1.1, 1.5)$, where no antenna samples exist, see hatched areas in Fig.~\ref{fig:ExC1}. Electrical length is an integer multiplicand of 0.1, the electrical size of one discretized pixel. The shortest antenna (in terms of $k \ell_y$) is a dipole, and the next shortest structure is either a driven dipole and a reflector or a driven dipole and a director with the director/reflector being generated in the closest distance to the driven dipole.

Comparing electrical length with regularity gives only one non-dominated solution, which corresponds to a dipole. All other solutions are electrically longer (in terms of $k \ell_y$) and/or have higher irregularity.

It is obvious from the bottom-right pane of Fig.~\ref{fig:ExC1} that higher realized gain requires a more sophisticated shape, which is reflected by higher irregularity~$R$.

Six different antenna samples are highlighted at the Pareto front in the top-right pane of Fig.~\ref{fig:ExC1} and correspondingly shown in Fig.~\ref{fig:ExC2}. It is seen that the Pareto front contains various shapes, from a simple dipole to arrangements with a driven dipole and reflectors/directors or highly irregular but valid solutions as the one shown in the bottom-right pane. The MOMA-AW algorithm found, on average, several hundreds of non-dominated solutions, one hundred of which are depicted in the supplementary material~\cite{KadlecCapek2024}.

\subsection{\RR{Realized Gain in Two Directions and Shape Regularity}}
\label{sec:ExD}

\RR{The final example utilizes arrangement from the previous Section~\ref{sec:ExC} which is shown in Fig.~\ref{fig:ExC3}. We determine the trade-off between realized gain pointing to $+y$ ($\vartheta=\pi/2$, $\varphi=\pi/2$) and $-y$ ($\vartheta=\pi/2$, $\varphi=3\pi/2$) directions with polarization $\UV{e} = \UV{\varphi}$, denoted for simplicity as $G_{\T{r}}(+y)$ and $G_{\T{r}}(-y)$, respectively. The objective function reads
\begin{equation}
    \widehat{f}_\T{D}(\srcRegion_n) = -w_1 G_{\T{r},n} (+y) - w_2 G_{\T{r},n}(-y) + w_3 R_n,
\end{equation}
with the realized gain defined in~\eqref{eq:refGain} and the regularity constraint in~\eqref{eq:exBreg}. The reference impedance is $Z_0 = 50\,\Omega$.}

\begin{figure}[t!]
    \centering
    \includegraphics[width=\columnwidth]{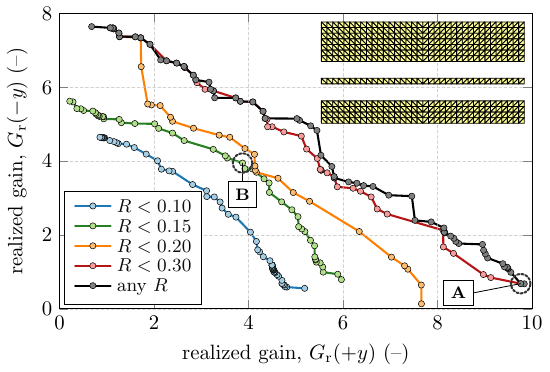}
    \caption{\RR{Trade-off between realized gain for two opposite directions $+y$ and $-y$, see coordinate system in Fig.~\ref{fig:ExC3}, with polarization $\UV{e} = \UV{\varphi}$. The approximate Pareto fronts are parameterized by regularity constraint~$R$, \cf{},~\eqref{eq:exBreg}. The Pareto front passing furthest to the right and at the top shows the best antenna performance if the regularity (point connections, isolated islands of material, etc.) are of no concern. The optimization mesh grid is shown as an inset in top right corner. Two solutions are chosen, denoted as A and B, and depicted in Fig.~\ref{fig:ExD2}.}}
    \label{fig:ExD1}
\end{figure}

\begin{figure*}[t!]
    \centering
    \includegraphics[width=1.9\columnwidth]{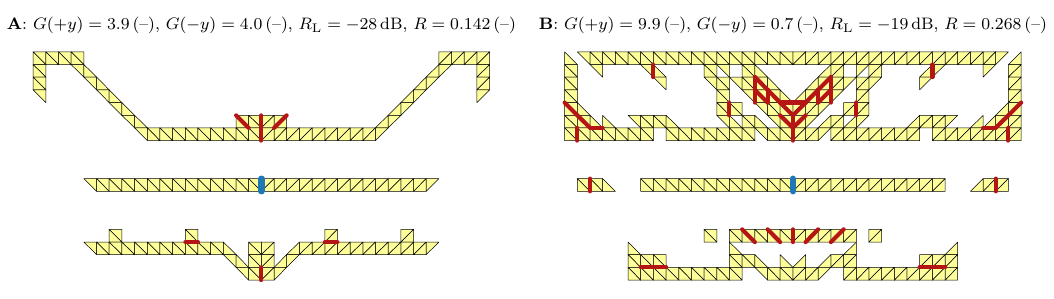}
    \caption{\RR{Antenna candidates picked from the Pareto front of Fig.~\ref{fig:ExD1}, labeled A and B. Discrete feeder is connected at the position highlighted by the thick blue edge. The performance is shown on the top. The left-right symmetry was enforced. The red edges mark locations with material discontinuity.}}
    \label{fig:ExD2}
\end{figure*}

\RR{The resulting trade-off is shown in Fig.~\ref{fig:ExD1}. The trade-off curves are parameterized with regularity constraint~$R$. Due to limited space, selected optimal shapes are shown in the supplementary material~\cite{KadlecCapek2024}, and only two candidates are depicted in Fig.~\ref{fig:ExD2}, including their performance shown in the title. It is seen that a higher value of regularity constraint leads to a more irregular shape with higher demands on post-processing (cutting slots, removing point connections). The traces for various regularity constraints become closer to the absolute Pareto front if the discretization of the optimization domain is further refined~\cite{KadlecCapek2024} (at the cost of the number of optimization unknowns). This is caused by the fact that the problematic features (point connections, infinitely thin slots, isolated islands of material) occur relatively less compared to the coarser mesh, however, they still appear. Therefore, it is advantageous to use the regularity constraint as one of the objective functions and obtain not only high-performance antennas but also shapes requiring less post-processing effort.}

\begin{figure}[t!]
    \centering
    \includegraphics[width=\columnwidth]{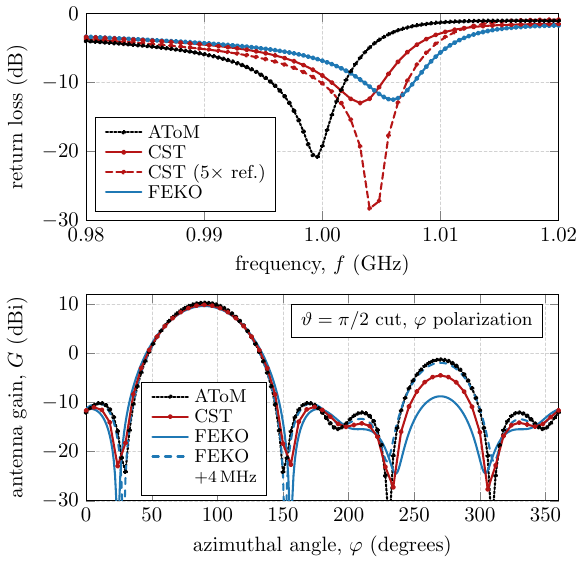}
    \caption{\RR{Comparison between performance of original model B from Fig.~\ref{fig:ExD2}, right pane, simulated in AToM~\cite{atom} with CST~\cite{cst} and FEKO~\cite{feko}. Top pane: return loss in decibels. The red dashed line is for a refined simulation grid in CST (250 triangles per wavelength). Bottom pane: far-field cut in dBi. The blue dashed line is a far-field cut in FEKO for frequency 1.004\,GHz.}}
    \label{fig:ExD4}
\end{figure}

\RR{The entire optimization procedure is verified by simulating the optimized shapes in commercial simulators CST Studio Suite~\cite{cst} and FEKO~\cite{feko}. We took shape~B from Fig.~\ref{fig:ExD2} since it contains more irregularities and, therefore, represents a more challenging case. The infinite slots shown by the red color are replaced by slots with finite width, taken as $P_x/20$, where $P_x = L_x/36$ is the width of one optimization pixel, see~\cite{Capeketal_ShapeSynthesisBasedOnTopologySensitivity} for further details regarding the role of the finite width of the slot. The point connections are removed by cutting out small boxes of vacuum. The models were discretized similarly as during the optimization (50 elements per wavelength), \ie{}, of the mesh density as shown in Fig.~\ref{fig:ExD2}. Notice that the used mesh density is considerably finer than the mesh obtained with default settings both in CST and FEKO.}

\RR{The return loss and far field cut in $\vartheta = \pi/2$ plane are shown in Fig.~\ref{fig:ExD4}. All models are reasonably matched, though, the performance can be further adjusted. To check the convergence with the mesh grid, the model in CST was refined to 250~elements per wavelength, see the red dashed line in the top pane of Fig.~\ref{fig:ExD4}. The resonant frequency is shifted by 0.45\%. The far-field cuts agree well. Similarly, as with the matching, the optimal performance might be slightly shifted, see the comparison between far-field cuts in FEKO for $1\,$GHz and $1.004\,$GHz. To conclude, the optimized structures can be translated into commercial solvers observing similar performance. Nevertheless, topology optimization is commonly considered the initial step in a design procedure, which helps to identify unknown (often non-intuitive) solutions. The initial design is adjusted afterward, see the role of topology optimization in structural engineering~\cite{Kazakis2017} or civil engineering~\cite{Baandrup2020}. As such, topology optimization at its current stage cannot fully replace antenna designers.}

\section{Possible Extensions and Discussion}
\label{sec:disc}

There are many possible extensions of the proposed algorithm that are beyond the scope of this paper.

The local algorithm is efficient in finding the local minima but consumes the majority of computational time. A possible remedy is to implement an adaptive mesh grid refinement based on a quad-tree algorithm compatible with the chosen basis functions.

The global algorithm (GA), in our case NSGA-II, serves one purpose only: to maintain diversity by the effective manipulation of candidate shapes identified by the local algorithm. Therefore, it can be replaced by a proprietary-designed scheme that would better exploit the full potential of the local rank-1 algorithm.

The proposed MO treatment is capable of determining a trade-off between the antenna's performance at different frequencies, \ie{},
\begin{equation}
    \widetilde{f} = \sum_i w_i f_i \left(\srcRegion_n , \omega_i \right), \quad \sum_i w_i = 1,
    \label{eq:add1}
\end{equation}
where~$\omega_i$ is angular frequency. The computational burden of~\eqref{eq:add1} scales linearly with the number of frequency points. This is, however, more than compensated by the increase in speed of the MOMA-AW algorithm, \ie{}, converting a series of SO problems (for each frequency) to MO problems (solving all frequency points at once).

The MOMA-AW algorithm is fast and efficient, capable of delivering high-quality, non-dominated sets of solutions, see \RR{Figs.~\ref{fig:boxplotA} and~\ref{fig:boxplotB}}. In light of this observation, this optimizer can be deployed as a powerful, robust data miner in physically informed machine learning~\cite{Karniadakis2021, Wu2024}.

Based on the recent success of artificial intelligence and data science, advanced techniques~\cite{Duda2000-fe} are available to analyze the results, either during or at the end of the optimization. A possible topic for future research is data clustering in the objective space to identify the general patterns and similarities of the antenna shapes in the variable space. For example, through the careful analysis of the non-dominated shapes and their performance depicted in Figs.~\ref{fig:ExB1} and~\ref{fig:ExC1}, see supplementary material~\cite{KadlecCapek2024} for details, it is seen that there is only a limited number of general patterns (a dipole, driven dipole and a reflector, a loop), each characterized by its average performance. Identifying these clusters and the main variations within them might help with surrogate modeling~\cite{Koziel2014-xd}.

\section{Conclusion}
\label{sec:conclu}

We have presented an adaptation of a single-objective memetic algorithm into its multi-objective variant. Instead of the common scalarization technique, which is time-consuming and frequently produces poor results, our proposal takes advantage of the existing rank-1 local algorithm, which produces locally optimal shapes, and combines it with the global heuristic algorithm that has been adopted in such a way so that each agent possesses its own set of adaptively updated weights.

The algorithm makes it possible to investigate various trade-offs, including electrical size or \RR{operating} frequency as optimization unknowns. For example, the designer obtains optimal antenna shapes for a range of predefined electrical sizes at once and can investigate how the optimal shape evolves with frequency.

The advantages of the proposed procedure are greater diversity and faster convergence, resulting in efficient optimization and a high-quality Pareto front. It has been demonstrated that the increase in computational speed is by several orders, as compared to the classical techniques, and scales up with the number of optimization unknowns, \ie{}, with more detailed models. It has also been shown that the algorithm generates uniform Pareto front sampling with considerably more solutions found than the state-of-the-art algorithms.

The features of the algorithm were verified on \RR{four} optimization problems involving two and three objectives. It has been confirmed that the bound on the Q factor is tight and scales as expected. The unpleasant trade-off between optimal bandwidth and matching of the electrical antenna was determined, noting that the manufacturability of the shape is an important aspect that influences performance. The last examples dealt with supergain, revealing scaling between the realized gain and the length of an \RR{antenna, or revealing trade-off between realized gain obtained for different directions.  The results show} that, in general, the more irregular shape can support a higher realized gain. The non-dominated solutions for all \RR{four} examples are provided in the supplementary material, including details regarding their performance.

There are many possible advancements and challenges for future research, namely to accelerate detailed models with adaptive multi-level, quad-tree mesh refinement, to improve the global algorithm to better suit the properties of rank-1 local updates, to consider further shape regularity constraints, to perform clustering and principal component analyses for the sets of non-dominated solutions, \RR{to incorporate the external archive strategy with comparison of previously tested combinations of agents and weights}, or to directly combine the algorithm with recent machine learning frameworks.

\appendices
\section{Update Strategy for Weighting Vectors}
\label{sec:algoWeightsUpdate}

The local neighborhood algorithm (see Algorithm~\ref{alg:locNeighborhoods}) is used to find out the so-called local neighborhood~$\M{L}_j$ that collects all shapes that are assigned to weighting vector~$\M{w}_j$ based on angle~$\nu_{i,j}$ between their~$\M{f}_i$ values and~$\M{w}_j$ (see Fig.~\ref{fig:WeightAssoc}). The procedure starts with a normalization of the objective function values as
\begin{equation}
    \widehat{{f}}_{m}  = \frac{f_m - z_{\mathrm{L}, m}}{z_{\mathrm{U}, m} - z_{\mathrm{L}, m} + 1},
    \label{eq:normObjF}
\end{equation}
where~$z_{\mathrm{L}, m}$ and~$z_{\mathrm{U}, m}$ denotes the~$m$-th value of the so-called utopian and nadir solution \cite{deb2002fast} (see Fig.~\ref{fig:WeightAssoc}). 

After local neighborhoods~$\M{L}_j$ are initialized as empty matrices for all weighting vectors, empty matrix~$\M{S}$, for storing solutions that will not enter the neighborhood of any weight, is created. The next step is to find the angular distance between solutions from normalized set~$\widehat{\M{F}}$ to all vectors from $\M{W}$. The angular distance between solution~$\widehat{\M{f}}_{i}$ and weighting vector~$\M{w}_j$ is given by:
\begin{equation}
    \nu_{i,j}  = \arccos\left(\frac{\M{w}_j^\trans \widehat{\M{f}}_i}{||\widehat{\M{f}}_i|| \, ||\M{w}_j||}\right),
    \label{eq:thetaAngle}
\end{equation}
where $|| \cdot ||$ is the L2 norm of a vector. Then, we find the closest weighting vector utilizing~$\nu$ for every solution from~$\widehat{\M{F}}$. This solution is assigned to the neighborhood of this weighting vector if its maximum capacity~$N_{\T{L}, \max}$ is not reached (see line 6 of Algorithm~\ref{alg:locNeighborhoods}). The size of the updated neighborhood denoted as $N_j$ is incremented by one. We set the value~$N_{\T{L}, \max} = 3$ to prevent many solutions from being allocated to a single weight. A solution with a minimal angle towards a weight with full neighborhood capacity is assigned to temporary set~$\M{S}$. 

\begin{algorithm}[t]
\footnotesize
\caption{New weighting vectors calculation.}
\label{alg:locNeighborhoods}
\hspace*{\algorithmicindent} \textbf{Input:} Matrices $\M{W}$, $\M{F}$, and parameters $N_{\T{L}}$, $N_{\T{W}}$, $\delta_{\mathrm{R}}$, $\delta_{\mathrm{C}}$. \\
\hspace*{\algorithmicindent} \textbf{Output:} Matrix of new weights $\M{W}_{\mathrm{new}}$ 
\begin{algorithmic}[1]
\State Find normalized solutions $\widehat{\M{F}}$ using \eqref{eq:normObjF}.
\State Initialize empty local neighborhoods $\M{L}$ and  matrix $\M{S}$.
\State Calculate distances $\nu_{i,j}$ between all $\widehat{\M{f}}_i \subset \widehat{\M{F}}$ and $\M{w}_j \subset \M{W}$.
\ForAll{$\widehat{\M{f}}_i \subset \widehat{\M{F}}$ }
    \State Find the closest $\V{w}_j$ according to $\nu_{i,j}$.
    \If{$N_j < N_{\T{L}, \max}$}
        \State Add $\widehat{\M{f}}_i$ to $\M{L}_j$
    \Else 
        \State Add $\widehat{\M{f}}_i$ to $\M{S}$
    \EndIf
\EndFor
\State Find the percentage of empty neighborhoods $p_0$.
\While{$p_0 > \delta_{\mathrm{C}}$}
    \State Place random solution from $\M{S}$ to some empty~$\M{L}_j$.
    \State Update $p_0$.
\EndWhile
\ForAll{$\widehat{\M{f}}_i \subset \M{S}$}
    \If{$r\in U\left(0,1\right) < \delta_{\mathrm{R}}$}
        \State Add $\widehat{\M{f}}_i$ to a random $\M{L}$ with minimal size. 
    \Else
        \State Add $\widehat{\M{f}}_i$ to a random $\M{L}$ with size $ > 0$.
    \EndIf
\EndFor
\State Initialize empty matrix $\M{W}_{\mathrm{new}}$.
\ForAll{$\M{w}_j $ such that $N_j == 0$}
    \State $\M{W}_{\mathrm{new}} = \M{W}_{\mathrm{new}} \cup \mathrm{WVG}\left(N_{\T{W}}, \,\M{w}_j,\,\xi\right)$
    \State Remove $\M{w}_j$ from $\M{W}$.
\EndFor
\State Return $\M{W}_{\mathrm{new}}$.
\end{algorithmic}
\end{algorithm}

After \RR{checking all solutions}, we allocate solutions from the temporary matrix $\M{S}$. If the total number of removed weighting vectors is smaller than parameter~$\delta_{\mathrm{C}}$, solutions from~$\M{S}$ are randomly redistributed to local neighborhoods that are empty. It enables the maintenance of weighting vectors that would have been deleted. Finally, if matrix~$\M{S}$ is still not empty, its members will be redistributed to local neighborhoods \RR{that have} a minimal number of members, different than zero. Again, exploration is preferred with this approach because it avoids creating crowded regions.

In the next step, we calculate the number of solutions~$N_j$ allocated to the neighborhood of every weighting vector~$\M{w}_j$. New weighting vectors are generated instead of~$N_0$ weighting vectors to which no solution has been assigned. The new weights are proposed by a weighting vector generator (WVG) introduced in \cite{reinaldo2021incorporation}. The procedure~$\T{WVG}\left(\M{w}_j,\, N_{\T{W}},\,\xi\right)$ proposes~$N_{\T{W}}$ new weights in the neighborhood of weight~$\M{w}_j$. According to recommendations published in~\cite{junqueira2022multi} we use value~$N_{\T{W}} = 3$ and we set the angle of aperture
\begin{equation}
    \xi  = \arccos\left(M^{-1/2}\right) \left(0.0353 M - 0.0322 \right),    
    \label{eq:apertureAngle}
\end{equation}
with $M$ being the number of objectives. 

All newly proposed weighting vectors are collected in matrix~$\M{W}_{\T{new}}$ (collecting column weighting vectors) which has size~$N_{\T{W}} N_0 \times N$. Then, $N_0$ weighting vectors from~$\M{W}_{\mathrm{new}}$ are selected to replace the weighting vectors with zero neighborhoods in~$\M{W}_t$ (\ie{}, no solution was assigned to them) so that the unit simplex in the objective space is covered as uniform as possible. The strategy is shown in Algorithm~\ref{alg:finalWeights} and it picks weighting vector~$\M{w}_j\subset \M{W}_{\T{new}}$ with the largest cumulative angular distance~$\sum_{i=1}^{N}\nu_{i,j}$ towards other column vectors of matrix~$\M{W}_t$.

\begin{algorithm}[t]
\footnotesize
\caption{Final weighting vectors choice.}
\label{alg:finalWeights}
\hspace*{\algorithmicindent} \textbf{Input:} Matrices $\M{W}_{\T{new}}$, $\M{W}_t$, and parameter $N_0$.\\
\hspace*{\algorithmicindent} \textbf{Output:} Matrix $\M{W}_t$. 
\begin{algorithmic}[1]
\While {$|\M{W}_t| < N_0$}
\State Find $\M{w}_j \subset \M{W}_{\T{new}}$ with max cumulative distance to all other vectors from~$\M{W}_t$.
\State Add $\M{w}_j$ to $\M{W}_t$.
\State Remove $\M{w}_j$ from $\M{W}_{\T{new}}$.
\EndWhile 
\State Return $\M{W}_t$.
\end{algorithmic}
\end{algorithm}

\section{\RR{Assignment of Weights to Solutions}}
\label{sec:algoWeights2Sol}

The weight assignment strategy is shown in Algorithm~\ref{alg:weightAssign}. It works on a set of available weights~$\M{W}_t$ and a set of new shapes~$\mathcal{O}_t$ represented by their individual objective function values~$\M{f} = \left[f_1, \dots, f_M\right]^\trans$. First, the minimization objective functions have to be multiplied by~$-1$ so that the objective function vectors correspond to the correct weighting vectors (see~\cite{junqueira2022multi} for a detailed explanation). All objective functions are normalized using~\eqref{eq:normObjF}. Second, we find angular distance matrix~$\M{A}$ between available weights and shapes using \eqref{eq:thetaAngle}. Then, we start a loop where~$k = \left\{1, \dots, N \right\}$ indices~$i$ and~$j$ of the minimal value in~$\M{A}$ are found in every iteration. Corresponding weight~$\M{w}_j$ is assigned to the shape with normalized~$\widehat{\M{f}}_i$. The distance matrix values for the $i$-th row and~$j$-th column are set to infinity not to assign them again.       

\begin{algorithm}[t]
\footnotesize
\caption{Weight-to-Solution Assignment Process.}
\label{alg:weightAssign}
\hspace*{\algorithmicindent} \textbf{Input:} Matrices $\M{F}$, $\M{W}$, and number of solutions $N$\\
\hspace*{\algorithmicindent} \textbf{Output:} Matrix $\M{W}_{\T{A}}$ 
\begin{algorithmic}[1]
\State Find normalized $\widehat{\M{f}}$ values for all $\M{f}_j \subset \M{F}$ \eqref{eq:normObjF}.
\State Find matrix  $A_{i,j} = \nu_{i,j}$ for all $\widehat{\M{f}}_i \subset \widehat{\M{F}}$ and $\M{w}_j \subset \M{W}$.
\For{$k=1$, $k\le N$, $k++$}. 
    \State Find indexes $i$ and $j$ where $\M{A}$ has the minimal value.
    \State Set $\M{w}_j$ to be $i$-th member of $\M{W}_{\T{A}}$.
    \State Set $\M{A}_{i,:} = \infty$ and $\M{A}_{:,j} = \infty$.
\EndFor
\State Return $\M{W}_{\T{A}}$.
\end{algorithmic}
\end{algorithm}

\section{Metrics for Multi-objective Optimization}
\label{sec:metricsMO}

The generational distance for found non-dominated set~$\mathcal{G}$ is computed according to \cite{Deb_MultiOOusingEA}
\begin{equation}
    \label{eq:GD}
    \mathrm{GD} = \dfrac{1}{|\mathcal{G}|} \left(\sum\limits_{i=1}^{|\mathcal{G}|} {d_i}^2\right)^{1/2},
\end{equation}
where~$|\mathcal{G}|$ is the size of set~$\OP{G}$, $d_i$ is the Euclidean distance measured in the objective space between solution~$\M{g}_i \subset \mathcal{G}$ and the nearest member of the true Pareto front~$\mathcal{G}^{\ast}$
\begin{equation}
    \label{eq:euclDist}
    d_i = \underset{k = 1}{\overset{|\mathcal{G}^{\ast}|}{\min}} \left(\sum_{m = 1}^{M} \left(f_m\left(\M{g}_i\right) - {f_m}\left({\M{g}_k}^{\ast}\right)\right)\right)^{1/2},
\end{equation}
where~$M$ is the number of objective functions and~$\M{g}^{\ast}$ is a member of the true Pareto front. The objective function values should be normalized using~\eqref{eq:normObjF} before~$\mathrm{GD}$ values are computed if there are huge differences between amplitudes of the objective function values. The~metric~$\mathrm{GD}$ expresses how close the found solutions are to the true Pareto front but does not indicate how well the entire Pareto front is covered. 

The metric hypervolume measures the coverage of the found Pareto front approximation. It is computed based on the formula~\cite{while2011fast}
\begin{equation}
    \label{eq:HV}
    \mathrm{HV} = \bigcup\limits_{i=1}^{|\mathcal{G}|}v\left(\gene_{i}\right),
\end{equation}
where~$v\left(\gene_{i}\right)$ is the volume of an $M$-dimensional hypercube with opposite corners at found solution $\gene_{i}$ and the nadir point~$\M{z}_{\mathrm{U}}$. If the true Pareto front is not known, the utopian and nadir solutions are determined from all the solutions found. The hypervolume metric indicates the quality of the Pareto front approximation in terms of the accuracy of individual solutions and coverage of the whole Pareto front.

\section{\RR{Distance Matrix --- Definition and Implementation}}
\label{sec:distMat}


Distance matrix~$\Dmat$ is a real-valued symmetric matrix containing positive numbers. It is defined elementwise as the maximum Euclidean distance
\begin{equation}
    \label{eq:Dmat}
    D_{pq} = \max \left\{ \left\| \V{r}_q - \V{r}_p \right\| \right\}, \quad \Dmat = [D_{pq}],
\end{equation}
where~$\V{r}_p \in A \left(\basisFcn_p\right)$, \ie{}, radius vector~$\V{r}_p$ spans a region where basis function~$\basisFcn_p$ lies, and similarly for vector~$\V{r}_q$. The evaluation of~\eqref{eq:Dmat} for RWG basis functions is to take the maximum distance from distances between all pairs of vertices of both~$p$-th and~$q$-th basis functions. 

To evaluate the smallest circumscribing sphere of an antenna~$\srcRegion_n$, the radius is found as
\begin{equation}
    \label{eq:arad}
    a_n = \dfrac{1}{2} \max \left\{ \T{diag} (\gene_n) \, \M{D} \, \T{diag} (\gene_n) \right\},
\end{equation}
where vector~$\gene_n$ fully characterizes the shape of antenna~$\srcRegion_n$ by identifying which degrees of freedom are enabled (material) or disabled (vacuum). Formula~\eqref{eq:arad} is easily approachable with logical indexing in, \eg{}, MATLAB or Python. Considering \mbox{rank-1} updates of the structure~\cite{Capeketal_InversionFreeEvaluationOfNearestNeighborsInMoM, 2021_capeketal_TSGAmemetics_Part1}, all possible radii can be calculated simultaneously at the cost of almost no computational burden.

The definition~\eqref{eq:Dmat} can be modified by replacing the radius vectors with their components only. Doing so, the meaning in~\eqref{eq:arad} is changed from calculating the smallest circumscribing sphere to, \eg{}, calculating the width/height of the antenna. Such a modification was used in Section~\ref{sec:ExC} to evaluate the electrical length of an \RR{antenna}.

\bibliographystyle{IEEEtran}

\begin{thebibliography}{10}
	\providecommand{\url}[1]{#1}
	\csname url@samestyle\endcsname
	\providecommand{\newblock}{\relax}
	\providecommand{\bibinfo}[2]{#2}
	\providecommand{\BIBentrySTDinterwordspacing}{\spaceskip=0pt\relax}
	\providecommand{\BIBentryALTinterwordstretchfactor}{4}
	\providecommand{\BIBentryALTinterwordspacing}{\spaceskip=\fontdimen2\font plus
		\BIBentryALTinterwordstretchfactor\fontdimen3\font minus
		\fontdimen4\font\relax}
	\providecommand{\BIBforeignlanguage}[2]{{%
			\expandafter\ifx\csname l@#1\endcsname\relax
			\typeout{** WARNING: IEEEtran.bst: No hyphenation pattern has been}%
			\typeout{** loaded for the language `#1'. Using the pattern for}%
			\typeout{** the default language instead.}%
			\else
			\language=\csname l@#1\endcsname
			\fi
			#2}}
	\providecommand{\BIBdecl}{\relax}
	\BIBdecl
	
	\bibitem{Balanis_Wiley_2005}
	C.~A. Balanis, \emph{Antenna Theory Analysis and Design}, 3rd~ed.\hskip 1em
	plus 0.5em minus 0.4em\relax Wiley, 2005.
	
	\bibitem{Fujimoto_Morishita_ModernSmallAntennas}
	K.~Fujimoto and H.~Morishita, \emph{Modern Small Antennas}.\hskip 1em plus
	0.5em minus 0.4em\relax Cambridge, Great Britain: Cambridge University Press,
	2013.
	
	\bibitem{VolakisChenFujimoto_SmallAntennas}
	J.~L. Volakis, C.~Chen, and K.~Fujimoto, \emph{Small Antennas: Miniaturization
		Techniques \& Applications}.\hskip 1em plus 0.5em minus 0.4em\relax
	McGraw-Hill, 2010.
	
	\bibitem{KozielOgurtsov_AntennaDesignBySimulationDrivenOptimization}
	S.~Koziel and S.~Ogurtsov, \emph{Antenna Design by Simulation-Driven
		Optimization}.\hskip 1em plus 0.5em minus 0.4em\relax Springer, 2014.
	
	\bibitem{Simon_EvolutionaryOptimizationAlgorithms}
	D.~Simon, \emph{Evolutionary Optimization Algorithms}.\hskip 1em plus 0.5em
	minus 0.4em\relax John Wiley \& Sons, 2013.
	
	\bibitem{BendsoeSigmund_TopologyOptimization}
	M.~P. Bendsoe and O.~Sigmund, \emph{Topology Optimization}, 2nd~ed.\hskip 1em
	plus 0.5em minus 0.4em\relax Berlin, Germany: Springer, 2004.
	
	\bibitem{Ohsaki_OptimizationOfFiniteDimensionalStructures}
	M.~Ohsaki, \emph{Optimization of Finite Dimensional Structures}.\hskip 1em plus
	0.5em minus 0.4em\relax CRC Press, 2011.
	
	\bibitem{JohnsonSamii1997_GAinEM}
	J.~M. Johnson and V.~Rahmat-Samii, ``Genetic algorithms in engineering
	electromagnetics,'' \emph{{IEEE} Antennas and Propagation Magazine}, vol.~39,
	no.~4, pp. 7--21, 1997.
	
	\bibitem{Haupt_Werner_GeneticAlgorithmsInEM}
	R.~L. Haupt and D.~H. Werner, \emph{Genetic Algorithms in
		Electromagnetics}.\hskip 1em plus 0.5em minus 0.4em\relax John Wiley \& Sons,
	2007.
	
	\bibitem{JorhsonRahmatSamii_GAandMoMforTheDesignOfIntegratedAntennas}
	J.~M. Johnson and Y.~Rahmat-Samii, ``Genetic algorithms and method of moments
	({GA/MOM}) for the design of integrated antennas,'' \emph{IEEE Trans.
		Antennas Propag.}, vol.~47, no.~10, pp. 1606--1614, Oct. 1999.
	
	\bibitem{Chen2016}
	Y.-S. Chen and Y.-H. Chiu, ``Application of multiobjective topology
	optimization to miniature ultrawideband antennas with enhanced pulse
	preservation,'' \emph{IEEE Antennas Wireless Propag. Lett.}, vol.~15, p.
	842–845, 2016.
	
	\bibitem{CismasuGustafsson_FBWbySimpleFreuqSimulation}
	M.~Cismasu and M.~Gustafsson, ``Antenna bandwidth optimization with single
	frequency simulation,'' \emph{IEEE Trans. Antennas Propag.}, vol.~62, no.~3,
	pp. 1304--1311, 2014.
	
	\bibitem{HassanWadbroBerggren_TopologyOptimizationOfMetallicAntennas}
	E.~Hassan, E.~Wadbro, and M.~Berggren, ``Topology optimization of metallic
	antennas,'' \emph{IEEE Trans. Antennas Propag.}, vol.~62, no.~5, pp.
	2488--2500, May 2014.
	
	\bibitem{2016_Liu_AMS}
	S.~Liu, Q.~Wang, and R.~Gao, ``{MoM}-based topology optimization method for
	planar metallic antenna design,'' \emph{Acta Mechanica Sinica}, vol.~32,
	no.~6, pp. 1058--1064, Dec. 2016.
	
	\bibitem{Tucek_etal_DensityBasedTOinMoMforQmin_2023}
	J.~Tucek, M.~Capek, L.~Jelinek, and O.~Sigmund, ``Density-based topology
	optimization in method of moments: Q-factor minimization,'' \emph{IEEE Trans.
		Antennas Propag.}, vol.~71, no.~12, p. 9738–9751, Dec. 2023.
	
	\bibitem{TucekEtAl_EuCAP2024_TopoOptGain}
	J.~Tucek, M.~Capek, and L.~Jelinek, ``Preliminary study on gain maximization
	via density-based topology optimization,'' in \emph{18th European Conference
		on Antennas and Propagation (EUCAP)}, Glasgow, Scotland, 2024.
	
	\bibitem{Capeketal_ShapeSynthesisBasedOnTopologySensitivity}
	M.~Capek, L.~Jelinek, and M.~Gustafsson, ``Shape synthesis based on topology
	sensitivity,'' \emph{IEEE Trans. Antennas Propag.}, vol.~67, no.~6, pp. 3889
	-- 3901, Jun. 2019.
	
	\bibitem{Jiang_etal_PixelAntenna_2022}
	F.~Jiang, S.~Shen, C.-Y. Chiu, Z.~Zhang, Y.~Zhang, Q.~S. Cheng, and R.~Murch,
	``Pixel antenna optimization based on perturbation sensitivity analysis,''
	\emph{IEEE Trans. Antennas Propag.}, vol.~70, no.~1, pp. 472--485, 2022.
	
	\bibitem{WangHum_APS_BinaryTopologyModel}
	Z.~Wang and S.~Hum, ``A new binary topology model for em surface unit cell
	design optimization,'' in \emph{2022 IEEE International Symposium on Antennas
		and Propagation and USNC-URSI Radio Science Meeting (AP-S/URSI)}, 2022, pp.
	67--68.
	
	\bibitem{2021_capeketal_TSGAmemetics_Part2}
	M.~Capek, M.~Gustafsson, L.~Jelinek, and P.~Kadlec, ``Optimal inverse design
	based on memetic algorithm -- {P}art {II}: {E}xamples and properties,''
	\emph{IEEE Trans. Antennas Propag.}, vol.~71, no.~11, pp. 8817--8829, 2024.
	
	\bibitem{2021_capeketal_TSGAmemetics_Part1}
	------, ``Optimal inverse design based on memetic algorithms -- {P}art {I}:
	{T}heory and implementation,'' \emph{IEEE Trans. Antennas Propag.}, vol.~71,
	no.~11, pp. 8806--8816, 2024.
	
	\bibitem{Deb_MultiOOusingEA}
	K.~Deb, \emph{Multi-Objective Optimization using Evolutionary
		Algorithms}.\hskip 1em plus 0.5em minus 0.4em\relax New York, United States:
	Wiley, 2001.
	
	\bibitem{Ehrgott_MulticriteriaOptimization}
	M.~Ehrgott, \emph{Multicriteria Optimization}.\hskip 1em plus 0.5em minus
	0.4em\relax Springer, 2005.
	
	\bibitem{hajela1992genetic}
	P.~Hajela and C.~Y. Lin, ``Genetic search strategies in multicriterion optimal
	design,'' \emph{Structural optimization}, vol.~4, pp. 99--107, 1992.
	
	\bibitem{murata1995moga}
	T.~Murata, H.~Ishibuchi \emph{et~al.}, ``{MOGA}: Multi-objective genetic
	algorithms,'' in \emph{IEEE International conference on evolutionary
		computation}, vol.~1.\hskip 1em plus 0.5em minus 0.4em\relax IEEE Piscataway,
	1995, pp. 289--294.
	
	\bibitem{ishibuchi1998multi}
	H.~Ishibuchi and T.~Murata, ``Multi-objective genetic local search for
	minimizing the number of fuzzy rules for pattern classification problems,''
	in \emph{1998 IEEE International Conference on Fuzzy Systems Proceedings.
		IEEE World Congress on Computational Intelligence (Cat. No. 98CH36228)},
	vol.~2.\hskip 1em plus 0.5em minus 0.4em\relax IEEE, 1998, pp. 1100--1105.
	
	\bibitem{parraga2017using}
	J.~Parraga-Alava, M.~Dorn, and M.~Inostroza-Ponta, ``Using local search
	strategies to improve the performance of {NSGA-II} for the multi-criteria
	minimum spanning tree problem,'' in \emph{2017 IEEE Congress on Evolutionary
		Computation (CEC)}.\hskip 1em plus 0.5em minus 0.4em\relax IEEE, 2017, pp.
	1119--1126.
	
	\bibitem{ma2019nsga}
	H.~Ma, A.~S. da~Silva, and W.~Kuang, ``{NSGA-II} with local search for
	multi-objective application deployment in multi-cloud,'' in \emph{2019 IEEE
		Congress on Evolutionary Computation (CEC)}.\hskip 1em plus 0.5em minus
	0.4em\relax IEEE, 2019, pp. 2800--2807.
	
	\bibitem{leung2020hybrid}
	M.-F. Leung, C.~A.~C. Coello, C.-C. Cheung, S.-C. Ng, and A.~K.-F. Lui, ``A
	hybrid leader selection strategy for many-objective particle swarm
	optimization,'' \emph{IEEE Access}, vol.~8, pp. 189\,527--189\,545, 2020.
	
	\bibitem{deb2002fast}
	K.~Deb, A.~Pratap, S.~Agarwal, and T.~Meyarivan, ``A fast and elitist
	multiobjective genetic algorithm: {NSGA-II},'' \emph{IEEE Transactions on
		Evolutionary Computation}, vol.~6, no.~2, pp. 182--197, 2002.
	
	\bibitem{trivedi2016survey}
	A.~Trivedi, D.~Srinivasan, K.~Sanyal, and A.~Ghosh, ``A survey of
	multiobjective evolutionary algorithms based on decomposition,'' \emph{IEEE
		Transactions on Evolutionary Computation}, vol.~21, no.~3, pp. 440--462,
	2016.
	
	\bibitem{junqueira2022multi}
	P.~P. Junqueira, I.~R. Meneghini, and F.~G. Guimar{\~a}es, ``Multi-objective
	evolutionary algorithm based on decomposition with an external archive and
	local-neighborhood based adaptation of weights,'' \emph{Swarm and
		Evolutionary Computation}, vol.~71, p. 101079, 2022.
	
	\bibitem{KadlecCapek2024}
	P.~Kadlec and M.~Capek, ``Multi-objective memetic algorithm with adaptive
	weights for inverse antenna design: Supplementary material,'' 2024,
	supplementary material.
	
	\bibitem{Hart_etal_RecentAdvancesInMemeticAlgorithms}
	W.~E. Hart, N.~Krasnogor, and J.~E. Smith, Eds., \emph{Recent Advances in
		Memetic Algorithms}.\hskip 1em plus 0.5em minus 0.4em\relax Springer, 2005.
	
	\bibitem{Harrington_FieldComputationByMoM}
	R.~F. Harrington, \emph{Field Computation by Moment Methods}.\hskip 1em plus
	0.5em minus 0.4em\relax Piscataway, New Jersey, United States: Wiley -- IEEE
	Press, 1993.
	
	\bibitem{das1998normal}
	I.~Das and J.~E. Dennis, ``Normal-boundary intersection: A new method for
	generating the pareto surface in nonlinear multicriteria optimization
	problems,'' \emph{SIAM journal on optimization}, vol.~8, no.~3, pp. 631--657,
	1998.
	
	\bibitem{marek2020fops}
	M.~Marek, P.~Kadlec, and M.~Capek, ``{FOPS}: A new framework for the
	optimization with variable number of dimensions,'' \emph{International
		Journal of RF and Microwave Computer-Aided Engineering}, vol.~30, no.~9, p.
	e22335, 2020.
	
	\bibitem{YaghjianBest_ImpedanceBandwidthAndQOfAntennas}
	A.~D. Yaghjian and S.~R. Best, ``Impedance, bandwidth and {Q} of antennas,''
	\emph{{IEEE} Trans. Antennas Propag.}, vol.~53, no.~4, pp. 1298--1324, Apr.
	2005.
	
	\bibitem{Chu_PhysicalLimitationsOfOmniDirectAntennas}
	L.~J. Chu, ``Physical limitations of omni-directional antennas,'' \emph{J.
		Appl. Phys.}, vol.~19, pp. 1163--1175, Dec. 1948.
	
	\bibitem{Capek_etal_2019_OptimalPlanarElectricDipoleAntennas}
	M.~Capek, L.~Jelinek, K.~Schab, M.~Gustafsson, B.~L.~G. Jonsson, F.~Ferrero,
	and C.~Ehrenborg, ``Optimal planar electric dipole antennas: Searching for
	antennas reaching the fundamental bounds on selected metrics,'' \emph{IEEE
		Antennas and Propagation Magazine}, vol.~61, no.~4, pp. 19--29, Aug. 2019.
	
	\bibitem{CapekGustafssonSchab_MinimizationOfAntennaQualityFactor}
	M.~Capek, M.~Gustafsson, and K.~Schab, ``Minimization of antenna quality
	factor,'' \emph{IEEE Trans. Antennas Propag.}, vol.~65, no.~8, pp.
	4115--4123, Aug, 2017.
	
	\bibitem{2018_Schab_Wsto}
	K.~Schab, L.~Jelinek, M.~Capek, C.~Ehrenborg, D.~Tayli, G.~A.~E. Vandenbosch,
	and M.~Gustafsson, ``Energy stored by radiating systems,'' \emph{IEEE
		Access}, vol.~6, pp. 10\,553--10\,568, 2018.
	
	\bibitem{Capeketal_InversionFreeEvaluationOfNearestNeighborsInMoM}
	M.~Capek, L.~Jelinek, and M.~Gustafsson, ``Inversion-free evaluation of nearest
	neighbors in method of moments,'' \emph{IEEE Antennas Wireless Propag.
		Lett.}, vol.~18, pp. 2311--2315, Apr. 2019.
	
	\bibitem{Best_ElectricallySmallResonantPlanarAntennas}
	S.~R. Best, ``Electrically small resonant planar antennas,'' \emph{IEEE
		Antennas Propag. Mag.}, vol.~57, no.~3, pp. 38--47, June 2015.
	
	\bibitem{RahmatSamii_Kovitz_Rajagopalan-NatureInspiredOptimizationTechniques}
	Y.~Rahmat-Samii, J.~M. Kovitz, and H.~Rajagopalan, ``Nature-inspired
	optimization techniques in communication antenna design,'' \emph{Proc. IEEE},
	vol. 100, no.~7, pp. 2132--2144, July 2012.
	
	\bibitem{YangAdams_SystematicShapeOptimizationOfSymmetricMIMOAntennasUsingCM}
	B.~Yang and J.~J. Adams, ``Systematic shape optimization of symmetric {MIMO}
	antennas using characteristic modes,'' \emph{{IEEE} Trans. Antennas Propag.},
	vol.~64, no.~7, pp. 2668--2678, July 2016.
	
	\bibitem{Capek_etal_RegularityConstraints_EuCAP2021}
	M.~Capek, V.~Neuman, J.~Tucek, L.~Jelinek, and M.~Gustafsson, ``Topology
	optimization of electrically small antennas with shape regularity
	constraints,'' in \emph{Proceedings of the 15th European Conference on
		Antennas and Propagation (EUCAP)}, 2021.
	
	\bibitem{Jayasinghe2015}
	J.~M. J.~W. Jayasinghe, J.~Anguera, D.~N. Uduwawala, and A.~Andújar,
	``Nonuniform overlapping method in designing microstrip patch antennas using
	genetic algorithm optimization,'' \emph{International Journal of Antennas and
		Propagation}, vol. 2015, p. 1–8, 2015.
	
	\bibitem{UzsokySolymar_TheoryOfSuperDirectiveLinearArrays}
	M.~Uzsoky and L.~Solym\'{a}r, ``Theory of super-directive linear arrays,''
	\emph{Acta Physica Academiae Scientiarum Hungaricae}, vol.~6, no.~2, pp.
	185--205, Dec. 1956.
	
	\bibitem{AltshulerODonnellYaghjianBest_AMonopoleSuperdirectiveArray}
	E.~E. Altshuler, T.~H. O'Donnell, A.~D. Yaghjian, and S.~R. Best, ``A monopole
	superdirective array,'' \emph{IEEE Trans. Antennas Propag.}, vol.~53, no.~8,
	pp. 2653--2661, Aug. 2005.
	
	\bibitem{Yaghjian_etal_RS2008}
	A.~D. Yaghjian, T.~H. O{\textquotesingle}Donnell, E.~E. Altshuler, and S.~R.
	Best, ``Electrically small supergain end-fire arrays,'' \emph{Radio Science},
	vol.~43, no.~3, pp. 1--13, May 2008.
	
	\bibitem{Debard2023}
	A.~Debard, A.~Clemente, A.~Tornese, and C.~Delaveaud, ``On the maximum end-fire
	directivity of compact antenna arrays based on electrical dipoles and huygens
	sources,'' \emph{IEEE Trans. Antennas Propag.}, vol.~71, no.~1, pp. 299--308,
	Jan. 2023.
	
	\bibitem{SchabEtAl_EigenvalueCrossingAvoidanceInCM}
	K.~R. Schab, J.~M. Outwater~Jr., M.~W. Young, and J.~T. Bernhard, ``Eigenvalue
	crossing avoidance in characteristic modes,'' \emph{IEEE Trans. Antennas
		Propag.}, vol.~64, no.~7, pp. 2617--2627, July 2016.
	
	\bibitem{Maseketal_ModalTrackingBasedOnGroupTheory}
	M.~Masek, M.~Capek, L.~Jelinek, and K.~Schab, ``Modal tracking based on group
	theory,'' \emph{IEEE Trans. Antennas Propag.}, vol.~68, no.~2, pp. 927--937,
	Feb. 2020.
	
	\bibitem{atom}
	\BIBentryALTinterwordspacing
	(2019) {A}ntenna {T}oolbox for {MATLAB} ({AToM}). Czech Technical University in
	Prague. {www.antennatoolbox.com}. [Online]. Available:
	\url{{www.antennatoolbox.com}}
	\BIBentrySTDinterwordspacing
	
	\bibitem{cst}
	\BIBentryALTinterwordspacing
	(2024) {CST Computer Simulation Technology}. Dassault Systemes. [Online].
	Available:
	\url{https://www.3ds.com/products-services/simulia/products/cst-studio-suite/}
	\BIBentrySTDinterwordspacing
	
	\bibitem{feko}
	\BIBentryALTinterwordspacing
	(2024) Altair {FEKO}. Altair. [Online]. Available:
	\url{https://www.altair.com/feko}
	\BIBentrySTDinterwordspacing
	
	\bibitem{Kazakis2017}
	G.~Kazakis, I.~Kanellopoulos, S.~Sotiropoulos, and N.~D. Lagaros, ``Topology
	optimization aided structural design: Interpretation, computational aspects
	and 3d printing,'' \emph{Heliyon}, vol.~3, no.~10, p. e00431, Oct. 2017.
	
	\bibitem{Baandrup2020}
	M.~Baandrup, O.~Sigmund, H.~Polk, and N.~Aage, ``Closing the gap towards
	super-long suspension bridges using computational morphogenesis,''
	\emph{Nature Communications}, vol.~11, no.~1, Jun. 2020.
	
	\bibitem{Karniadakis2021}
	G.~E. Karniadakis, I.~G. Kevrekidis, L.~Lu, P.~Perdikaris, S.~Wang, and
	L.~Yang, ``Physics-informed machine learning,'' \emph{Nature Reviews
		Physics}, vol.~3, no.~6, p. 422–440, May 2021.
	
	\bibitem{Wu2024}
	Q.~Wu, W.~Chen, C.~Yu, H.~Wang, and W.~Hong, ``Machine-learning-assisted
	optimization for antenna geometry design,'' \emph{IEEE Trans. Antennas
		Propag.}, vol.~72, no.~3, p. 2083–2095, Mar. 2024.
	
	\bibitem{Duda2000-fe}
	R.~O. Duda, P.~E. Hart, and D.~G. Stork, \emph{\BIBforeignlanguage{en}{Pattern
			Classification}}, 2nd~ed., ser. A Wiley-Interscience publication.\hskip 1em
	plus 0.5em minus 0.4em\relax Nashville, TN: John Wiley \& Sons, Oct. 2000.
	
	\bibitem{Koziel2014-xd}
	S.~Koziel and S.~Ogurtsov, \emph{\BIBforeignlanguage{en}{Antenna design by
			simulation-driven optimization}}, 2014th~ed., ser. SpringerBriefs in
	Optimization.\hskip 1em plus 0.5em minus 0.4em\relax Cham, Switzerland:
	Springer International Publishing, Feb. 2014.
	
	\bibitem{reinaldo2021incorporation}
	I.~Reinaldo~Meneghini, F.~Gadelha~Guimar{\~a}es, A.~Gaspar-Cunha, and
	M.~Weiss~Cohen, ``Incorporation of region of interest in a
	decomposition-based multi-objective evolutionary algorithm,'' \emph{Advances
		in Evolutionary and Deterministic Methods for Design, Optimization and
		Control in Engineering and Sciences}, pp. 35--50, 2021.
	
	\bibitem{while2011fast}
	L.~While, L.~Bradstreet, and L.~Barone, ``A fast way of calculating exact
	hypervolumes,'' \emph{IEEE Transactions on Evolutionary Computation},
	vol.~16, no.~1, pp. 86--95, 2011.
	
\end{thebibliography}

\end{document}